\documentclass{article}

\PassOptionsToPackage{numbers, compress}{natbib}

\usepackage[eandd, final]{neurips_2026}

\usepackage[utf8]{inputenc} 
\usepackage[T1]{fontenc}    
\usepackage{hyperref}       
\usepackage{url}            
\usepackage{booktabs}       
\usepackage{amsfonts}       
\usepackage{nicefrac}       
\usepackage{microtype}      
\usepackage{xcolor}         
\definecolor{darkgreen}{rgb}{0.0, 0.45, 0.0}

\usepackage{float}
\usepackage{placeins}       
\usepackage{amsmath}
\usepackage{graphicx}
\usepackage{longtable}
\usepackage{amssymb}

\title{MulTaBench: Benchmarking Multimodal Tabular Learning with Text and Image}

\author{
  \textbf{Alan Arazi}$^{*,1,2}$ \quad
  \textbf{Eilam Shapira}$^{*,1}$ \quad
  \textbf{Shoham Grunblat}$^{1}$ \quad
  \textbf{Mor Ventura}$^{1}$ \\
  \textbf{Elad Hoffer}$^{3}$ \quad
  \textbf{Gioia Blayer}$^{4}$ \quad
  \textbf{David Holzmüller}$^{4}$ \quad
  \textbf{Lennart Purucker}$^{2,5}$ \\
  \textbf{Gaël Varoquaux}$^{4,6}$ \quad
  \textbf{Frank Hutter}$^{2,7,5}$ \quad
  \textbf{Roi Reichart}$^{1}$ \\
  \vspace{0.5em} \\
  \small $^*$Equal contribution \\
  \small $^{1}$Technion -- Israel Institute of Technology \quad  $^{2}$Prior Labs \quad
  $^{3}$NVIDIA \\
  \small $^{4}$SODA Team, INRIA Saclay, Palaiseau \quad 
  $^{5}$University of Freiburg \quad
  $^{6}$Probabl \quad
  $^{7}$ELLIS Institute Tübingen\\
  \texttt{\{alanarazi7, eilam.shapira, roireichart\}@gmail.com}
}

\begin{document}

\maketitle

\begin{abstract}
Tabular Foundation Models have recently established the state of the art in supervised tabular learning, by leveraging pretraining to learn generalizable representations of numerical and categorical structured data. However, they lack native support for unstructured modalities such as text and image, and rely on frozen, pretrained embeddings to process them. On established Multimodal Tabular Learning benchmarks, we show that tuning the embeddings to the task improves performance. Existing benchmarks, however, often focus on the mere co-occurrence of modalities; this leads to high variance across datasets and masks the benefits of task-specific tuning. To address this gap, we introduce MulTaBench, a benchmark of 40 datasets, split equally between image-tabular and text-tabular tasks. We focus on predictive tasks where the modalities provide complementary predictive signal, and where generic embeddings lose critical information, necessitating Target-Aware Representations that are aligned with the task. Our experimental results demonstrate that the gains from target-aware representation tuning generalize across both text and image modalities, several tabular learners, encoder scales, and embedding dimensions. MulTaBench constitutes the largest image-tabular benchmarking effort to date, spanning high-impact domains such as healthcare and e-commerce. It is designed to enable the research of novel architectures which incorporate joint modeling and target-aware representations, paving the way for the development of novel Multimodal Tabular Foundation Models.\footnote{\url{https://github.com/alanarazi7/MulTaBench}.}
\end{abstract}

\section{Introduction}
\label{sec:introduction}

Tabular Foundation Models (TFMs) \cite{van_breugel_position_2024, hollmann_tabpfn_2022, hollmann_accurate_2025, qu_tabicl_2025, grinsztajn_tabpfn-25_2026, qu_tabiclv2_2026} have recently emerged as the state of the art (SOTA) for supervised tabular learning \cite{erickson_tabarena_2025, ye_closer_2025}. They have surpassed gradient-boosted decision trees (GBDTs) \cite{breiman_random_2001, chen_xgboost_2016, ke_lightgbm_2017, prokhorenkova_catboost_2018}, which have historically been the leading approach \cite{shwartz-ziv_tabular_2022, grinsztajn_why_2022, mcelfresh_when_2023}. Recently, these versatile learners have been extended to causal inference \cite{robertson_-pfn_2025}, graph learning \cite{hayler_bringing_2025}, and time-series \cite{hoo2024tabular}. However, the best-performing TFMs \cite{grinsztajn_tabpfn-25_2026, qu_tabiclv2_2026} are trained exclusively on structured numerical data, making them fundamentally unimodal: unstructured inputs must be preprocessed via external embedding models \cite{wang_text_2024, simeoni_dinov3_2025}, with no unified support for modalities such as text and image.

Yet, in many high-impact domains, tabular problems are multimodal: e-commerce listings \cite{das2024towards, sukel2024multimodal, shapira2024can}, social media feeds \cite{meghawat2018multimodal, ophir2020deep, badian2023social}, and medical health records \cite{huang_fusion_2020, cui_deep_2023, duenias_hyperfusion_2025, fu_unleashing_2025} combine image and text with numerical features. While early work has begun extending TFMs to integrate text \cite{arazi_tabstar_2025, spinaci_contexttab_2025}, these extensions often compromise the model's core tabular performance, and inherent support for visual modalities remains entirely absent. One might turn to Large Language and Vision-Language Models (LLMs/VLMs), which natively process unstructured inputs, but they are not suited for the inductive biases of tabular data; specifically, they are unoptimized for the relational structure \cite{fang_large_2024} and are suboptimal for numerical features \cite{van_breugel_position_2024}. Addressing these limitations requires architectures that combine the numerical precision of TFMs while maintaining the rich input handling of multimodal foundation models. However, evaluating such a unified approach is difficult because the diverse nature of tasks within Multimodal Tabular Learning (MMTL) \cite{jiang_representation_2026, kim_multimodalpfn_2025} is not yet fully characterized; existing benchmarks \cite{shi_benchmarking_2021, lu_mug_2023, kim_carte_2024, tang_bag_2024, mraz_towards_2025} primarily highlight the coexistence of modalities, unintentionally grouping together problems that require fundamentally different modeling solutions.

\begin{figure}[!t]
\centering
\includegraphics[width=0.93\linewidth]{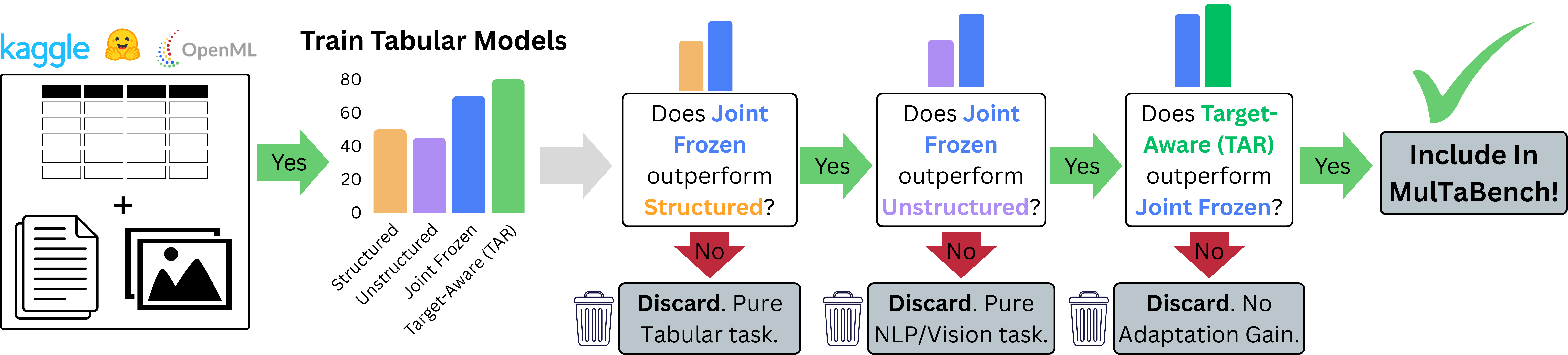}
\caption{The MulTaBench Curation Pipeline. Datasets are included if joint prediction outperforms unimodal baselines and if Target-Aware Representations improve on frozen, off-the-shelf embeddings.}
\label{fig:curation_flow}
\end{figure}

To characterize these problems, we observe that tabular models require inputs to be represented as feature columns, so high-dimensional images and texts must be compressed into compact representations. Consequently, embeddings act as lossy summaries, as they capture only a fraction of the raw input's information by design \cite{weller_theoretical_2025}. In order to generalize well, pretrained embedding models are optimized for broad semantic content, such as distinguishing an X-ray from a mammogram, at the expense of fine-grained details like precise size estimations or localized anomalies  \cite{pantazopoulos_lost_2024, li_lost_2025}. While this compression is effective for global semantic mapping, it fails to preserve the specialized signals required for fine-grained MMTL tasks. For example, the optimal representation of a chest X-ray differs depending on whether the tabular task is to diagnose pneumonia or a rib fracture, and whether the patient is a young athlete or an elderly smoker. We thus advocate for the need for Target-Aware Representations (TAR): embeddings that are tuned to the target and, ideally, to the other modalities.

Consider, for example, the task of pneumonia detection from a patient record combining age and smoking status with chest X-ray images. We argue that to study MMTL, a dataset should satisfy two properties: (1) \textit{Joint Signal}, where each modality provides complementary information that contributes to the overall predictive performance, and (2) \textit{Task-awareness}, where task-agnostic representations fail to capture the details required for a given objective. In our example, both the X-ray and the clinical profile offer unique, complementary information, and steering the image embedding to detect subtle signs of inflammation in the lungs should improve diagnostic accuracy.

To translate these theoretical properties into a measurable test, we develop an algorithmic pipeline that quantifies whether a dataset complies with the aforementioned requirements. This approach approximates these properties by evaluating each task across a broad suite of tabular learners, ranging from light GBDTs to SOTA TFMs. 
To evaluate for \textit{Joint Signal}, we demand a performance drop when either modality is removed, verifying that each input strengthens the predictive power. For \textit{Task-awareness}, we finetune the encoder's last 3 layers with LoRA \cite{hu_lora_2021} on the prediction target as a preprocessing step, and we expect these representations to outperform frozen ones when passed to tabular models. Crucially, our experiments confirm that target-aware representations outperform frozen embeddings across established MMTL benchmarks; however, we find that the magnitude of these gains is highly dataset-dependent, suggesting they represent distinct classes of MMTL tasks.

Building on this framework, we introduce \textbf{MulTaBench}, a benchmark of 40 datasets balanced between image-tabular and text-tabular tasks, as well as classification and regression objectives. To ensure a comprehensive evaluation, the benchmark incorporates a wide range of sample sizes and feature counts, while spanning a diverse set of domains to capture the heterogeneity of real-world multimodal tabular data. MulTaBench represents the largest image-tabular benchmarking effort to date, and the first MMTL benchmark to explicitly test the limits of frozen embeddings. Rather than attempting to cover the entire MMTL space, it specifically isolates datasets that benefit from joint fusion and favor target-aware adaptation. We additionally release 40 candidate datasets that satisfy the \textit{Joint Signal} criterion alone, extending MulTaBench to an 80-dataset collection. This allows models to be evaluated both where target-awareness is essential and where standard frozen embeddings suffice. Demonstrating the robustness of our curation criteria, we show that the gains from target-aware tuning generalize consistently across a diverse suite of independent tabular learners, encoder scales, and embedding dimensions. These findings suggest that designing novel architectures which contextualize the representations of unstructured modalities can push the boundaries of MMTL, and we believe that MulTaBench will be instrumental for developing true Multimodal TFMs.

\section{Related Work}
\label{sec:related}

\paragraph{Tabular Foundation Models.} The landscape of tabular learning shifted with Prior-data Fitted Networks (PFNs) \cite{muller_transformers_2021}, which pretrain transformers over synthetic tabular datasets with in-context learning (ICL) \cite{brown_language_2020}. The TabPFN family \cite{hollmann_tabpfn_2022, hollmann_accurate_2025, grinsztajn_tabpfn-25_2026, garg_real-tabpfn_2025} pioneered this direction. Multiple subsequent works \cite{qu_tabicl_2025, qu_tabiclv2_2026, ma_tabdpt_2025, zhang_mitra_2025, spinaci_contexttab_2025, zhang_limix_2025, bouadi_orion-msp_2025} advanced the paradigm with improvements spanning synthetic data diversity, real-world data pretraining, and architectural scalability. Among these, ConTextTab \cite{spinaci_contexttab_2025} is the only PFN to incorporate textual fields, yet it does not process raw strings; instead, it relies on external, frozen text embeddings as static inputs, decoupling the representation from the tabular learning objective. In addition, several non-PFN approaches \cite{yan_making_2023, kim_carte_2024, kim_table_2025} also incorporate semantic awareness, but likewise treat text representations as frozen. TabSTAR \cite{arazi_tabstar_2025} represents a fundamental shift: rather than processing fixed representations, it jointly trains both the textual and tabular encoders, successfully demonstrating that TAR are essential for MMTL. However, it lacks support for images and its non-ICL architecture compromises its numerical performance.

\paragraph{LLMs and VLMs.} Recent years have seen the rise of LLMs and their evolution into VLMs \cite{wu_multimodal_2023, yin_survey_2024, caffagni_revolution_2024}. These powerful models \cite{singh_openai_2025, comanici_gemini_2025} typically employ a unified transformer architecture \cite{vaswani_attention_2017} to process interleaved modalities within a single sequence, offering a path to integrate tabular data with text and image; however, research has primarily focused on text-tabular tasks \cite{fang_large_2024}. TabLLM \cite{hegselmann_tabllm_2023} explored different strategies to serialize the tabular data into natural language, and TabuLa-8B \cite{gardner_large_2024} and TabGemma \cite{schindler_tabgemma_2025} combined continued pretraining of LLMs on tabular corpora \cite{eggert_tablib_2023} with architectural modifications, achieving strong few-shot performance. Nevertheless, the autoregressive nature of LLMs is misaligned with the structure of tabular data, and their tokenization process damages numerical precision \cite{thawani_representing_2021, spathis_first_2024}. Furthermore, their massive scale introduces prohibitive costs for high-throughput inference, while their extensive pretraining risks memorizing evaluation data \cite{bordt_elephants_2024, gorla_illusion_2026}. Consequently, generative architectures remain largely impractical for discriminative MMTL.

\paragraph{Joint Multimodal Tabular Learning Architectures.} Despite various proposals to advance MMTL \cite{hager_best_2023, jiang_tabular_2024, ebrahimi_lanistr_2024, hu_pytorch_2024, leonardis_tip_2025, azizi26a}, the field still lacks a true multimodal foundation model for tabular data with text and images. AutoML \cite{he_automl_2021} frameworks \cite{shi_benchmarking_2021, tang_autogluon-multimodal_2024, tang_bag_2024}, led by AutoGluon-Multimodal \cite{tang_autogluon-multimodal_2024}, demonstrated the benefit of joint modeling by combining tabular, text and image encoders. However, their reliance on a non-ICL transformer \cite{gorishniy_revisiting_2021} as the tabular backbone limits their tabular capacities. Similarly, TabSTAR \cite{arazi_tabstar_2025} introduced a jointly pretrained text-tabular architecture and achieved strong performance on text-tabular classification tasks, but it struggled with regression tasks and with unimodal tabular benchmarks \cite{erickson_tabarena_2025}. Recent attempts have built on stronger tabular foundations, by expanding the PFN paradigm with multimodal fusion strategies. TIME \cite{luo_time_2025} proposed a late-fusion approach in an image-tabular setup, but missed cross-modal interactions and achieved mixed results when employing finetuning. MultiModalPFN \cite{kim_multimodalpfn_2025} fused TabPFN with visual and textual backbones, but assumed frozen multimodal embeddings. To conclude, no existing model has successfully maintained SOTA performance on tabular tasks while learning TAR for text and images.

\paragraph{Text-Tabular Benchmarks.} Existing text-tabular benchmarks differ significantly in their curation philosophy and dataset scale. The Multimodal AutoML Benchmark \cite{shi_benchmarking_2021} introduced 18 datasets with deliberate diversity in task type and predictive signal. \citet{grinsztajn_vectorizing_2023} filtered 14 datasets from a bigger pool, where the text features provided a significant gain over a numerical-only baseline. TextTabBench \cite{mraz_towards_2025} curated 13 text-tabular datasets, focusing on longer text fields while ensuring both the text modality and numerical features contribute to the prediction. CARTE \citep{kim_carte_2024} collected 51 datasets, mainly featuring short strings and high-cardinality categories, typically present in knowledge graphs. While these efforts were instrumental in advancing research on tabular data with strings, none of them were deliberately designed to isolate tasks where static representations fail to capture the necessary predictive signal. This limitation extends to concurrent works such as STRABLE \cite{blayer2026strable}, which benchmarks tabular learning with string columns, and BeyondArena \cite{purucker2026beyond}, which includes a subset of text-tabular datasets. Importantly, as we show in \S~\ref{par:text_tabular_curation}, most of the datasets included in the aforementioned benchmarks do not pass our curation pipeline. Consequently, potential performance gains that native Multimodal TFMs are designed to deliver might be overlooked. For example, ConTextTab set the SOTA for the CARTE benchmark \cite{spinaci_contexttab_2025}, but struggles on MulTaBench (see \S~\ref{sec:results}).

\paragraph{Image-Tabular Benchmarks.} The availability of image-tabular benchmarks is highly limited. MuG \citep{lu_mug_2023} introduced 4 data sources from the gaming domain combining tabular data with text and image, but offering limited domain diversity. Similarly, \citet{tang_bag_2024} curated 11 tabular datasets with images, but without quantifying the image signal's necessity. In concurrent work, VT-Bench \cite{jia2026vt} curates 14 visual-tabular datasets to study fusion methods, rather than filtering for datasets where static representations fail. As detailed in \S~\ref{par:image_tabular_curation}, these datasets often fail our curation pipeline, and, in certain cases, suffer from additional quality issues. This historical lack of large, accessible benchmarks led recent work such as TIME \cite{luo_time_2025} and MultimodalTabPFN \cite{kim_multimodalpfn_2025}, to rely on a self-selected group of datasets, limiting the generalizability of their findings. We address this gap by doubling the benchmark size and assuring that the image representations are central for MMTL.

\paragraph{Limits of Frozen Representations.} 
Pretrained representations are optimized for general-purpose objectives and often fail to capture the fine-grained, task-specific details necessary for downstream performance \cite{tong_eyes_2024, liu_data_2025, gisserot-boukhlef_should_2025, cao_tipsv2_2026}. \citet{weller_theoretical_2025} provide a theoretical basis for this limitation, demonstrating how RAG systems \cite{lewis_retrieval-augmented_nodate} that rely on static embeddings can fail on even seemingly simple cases. To overcome this problem, alternative approaches \cite{khattab_colbert_2020, malaviya_quest_2023, fan_survey_2024, tang_we_2024, edge_local_2025, wang_jina-reranker-v3_2025, pu_customized_2025, koshorek_structured_2025, hoffer2026retrievalwithinintrinsiccapability} enabled the contextualization of document representations in the presence of the query. Similar limitations were also illustrated in VQA \cite{antol_vqa_2015}, where encoding images independently of the question leads to information loss, as the query determines which image regions are predictive \cite{ganz_question_2024, li_lost_2025}. To overcome these limitations, VLMs have evolved toward deep multimodal alignment \cite{radford_learning_2021, alayrac_flamingo_2022, liu_visual_2023}, and we argue that MMTL should undergo a similar evolution, moving away from decoupled preprocessing and frozen embeddings in favor of a joint learning approach.

\section{Benchmarking Multimodal Tabular Learning}\label{sec:benchmarking}

MMTL \cite{jiang_representation_2026, kim_multimodalpfn_2025} refers to prediction tasks where inputs combine structured data, such as numerical and categorical columns, with unstructured modalities like text or image. Within each modality, a dataset may contain multiple features, such as various numerical columns or distinct text fields. For the clarity of analysis in this section, we assume that a single unstructured modality is paired with the tabular data. However, our logic naturally extends to trimodal datasets, as discussed in \S\ref{par:trimodal_datasets}.

\subsection{Desiderata for Multimodal Tabular Learning datasets}
\label{sec:benchmark_desiderata}

Consider a pneumonia dataset where each observation pairs structured clinical metadata, such as age and smoking status, with textual clinical notes or chest X-ray images to predict diagnosis.
While this seems a natural candidate for an MMTL benchmark, we argue that whether this dataset represents a challenging MMTL problem depends on two properties that must hold:

\paragraph{Joint Signal.} Following the principle in \citet{mraz_towards_2025}, we require each modality to carry independent signal about the target, so the joint predictive performance exceeds the union of unimodal performances. In the pneumonia case, the X-ray encodes spatial lung patterns, while age and smoking status convey clinical risk factors that provide information invisible in pixels. This criterion could optionally capture cross-modal interactions, where one modality might only become discriminative once conditioned on the other. For instance, increased reticular markings may signal acute infection in non-smokers, yet merely represent baseline chronic changes in a long-term smoker; the visual feature only becomes discriminative when conditioned on the tabular history. A modality can fail this criterion if it carries no signal (e.g., a clinical note containing only administrative metadata), or if its signal is already captured by another modality and thus provides no predictive gain (e.g., a note that merely transcribes the patient's age and smoking status, which already exist as structured features).

\paragraph{Task-awareness} We define \textit{Task-awareness} as a property of the computational problem where the optimal representation of an unstructured modality depends on the task context. A task exhibits \textit{Task-awareness} when the predictive signal is latent in the raw input at a level of granularity that differs from the modality's global semantic meaning. Because general-purpose encoders are optimized to preserve high-level properties while discarding low-level variance, such as exact wording \cite{weller_theoretical_2025} or fine-grained spatial textures \cite{pantazopoulos_lost_2024}, they often discard the specific nuances required for MMTL. Recovering this signal necessitates TAR, which steer the representation to focus on the details relevant to the specific target.\footnote{While joint tuning with structured features could add predictive value, explicitly requiring it would be unnecessarily strict.} In our pneumonia example, a generic model might identify the scan's global anatomy, whereas TAR would preserve the tiny visual patterns in the lung tissue that are key for diagnosis. Conversely, a task lacks \textit{Task-awareness} if the predictive signal is coarse enough to be captured by task-agnostic embeddings; for instance, if the objective is simply to categorize the scan type rather than identify a specific pathology, TAR would provide no significant advantage. 

\subsection{The Curation Pipeline}
\label{sec:benchmark_design}

To bridge the gap between the theoretical desiderata and the empirical curation, we establish an evaluation protocol based on 4 experimental conditions, as summarized in Table~\ref{tab:conditions} and Figure~\ref{fig:curation_flow}. The conditions vary by the features included and the specific representation of the unstructured modalities.

\begin{table}[ht]
\centering
\caption{Experimental Conditions. Breakdown by feature composition and representation strategy.}
\label{tab:conditions}
\begin{tabular}{lccc}
\hline
\textbf{Condition} & \textbf{Structured} & \textbf{Unstructured} & \textbf{Target-Aware (TAR)} \\ \hline
Unimodal Structured   & $\checkmark$ & $\times$     & -- \\
Unimodal Unstructured & $\times$     & $\checkmark$ & $\times$ \\
Joint Frozen          & $\checkmark$ & $\checkmark$ & $\times$ \\
Joint TAR             & $\checkmark$ & $\checkmark$ & $\checkmark$ \\ \hline
\end{tabular}
\end{table}

Our approach intentionally entangles task properties with algorithmic solutions in order to isolate datasets that align with our criteria and that current models struggle with.  Embeddings are extracted using \textit{e5-v2-small} \cite{wang_text_2024} for texts and \textit{DINO-v3-small} \cite{simeoni_dinov3_2025} for images, selected for their high performance-to-parameter efficiency \cite{muennighoff_mteb_2023}. To implement our proposed TAR condition, we finetune the last 3 layers on the prediction target using LoRA \cite{hu_lora_2021}. Crucially, this adaptation is performed as a specialized preprocessing step without the structured features and shared across learners. Representations are down-projected with PCA \cite{mackiewicz_principal_1993} to a dimension of 30, to ensure computational efficiency. We employ 5 diverse tabular learners: GBDTs (LightGBM \cite{ke_lightgbm_2017} and CatBoost \cite{prokhorenkova_catboost_2018}), the MLP-based TabM \cite{gorishniy_tabm_2025}, and the TFMs TabPFNv2 \cite{hollmann_accurate_2025} and TabPFN-2.5 \cite{grinsztajn_tabpfn-25_2026}. For each candidate dataset, we evaluate every model in each condition over 5 random seeds, subsampling up to 10,000 examples per run for cost-effectiveness. Our metric is AUC for classification tasks and $R^2$ for regression tasks. 

\begin{figure}[h]
\centering
\includegraphics[width=0.98\textwidth]{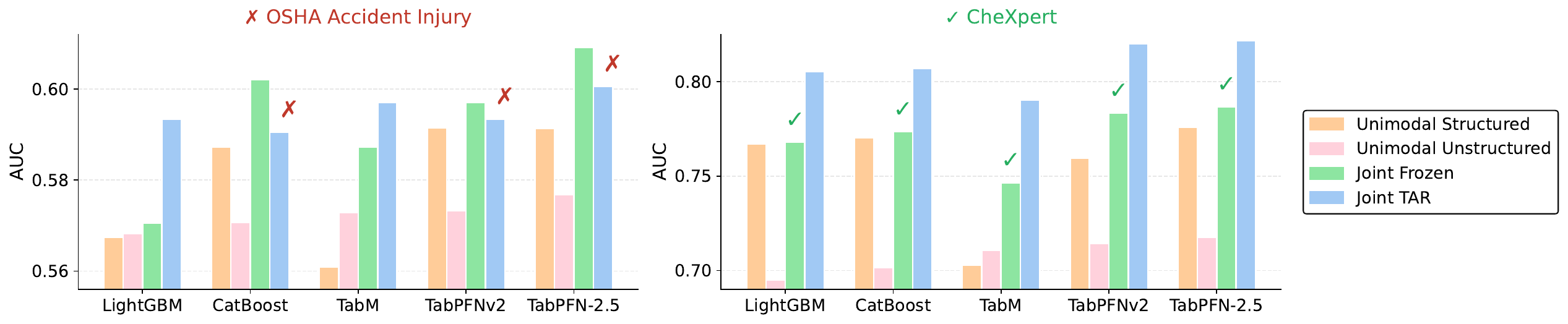}
\caption{Curation protocol over candidate datasets. Mean AUC per model and condition. The \textit{OSHA Accident Injury} dataset is rejected as \textit{TAR} fails to consistently improve over \textit{Joint Frozen}.}
\label{fig:curation_example}
\end{figure}

\paragraph{Acceptance Criteria.} To pass the curation filter, a dataset should satisfy two conditions across at least 3 out of 5 learners: (1) For \textit{Joint Signal}, performance over the \textit{Joint Frozen} condition should be higher than both \textit{Unimodal Structured} and \textit{Unimodal Unstructured} variants. This ensures that the unstructured modality is relevant, while also prevents the dataset from collapsing into a pure Natural Language Processing or Computer Vision task; and (2) For \textit{Task-awareness} we require that the \textit{Joint TAR} condition will improve performance over the \textit{Joint Frozen} condition, isolating the gain from representation tuning. Figure~\ref{fig:curation_example} illustrates the protocol over concrete examples. \S\ref{sec:results} analyzes how sensitive the final benchmark is to the acceptance criteria and to the choice of curation learners, and Appendix~\ref{app:curation} provides a formal definition of the acceptance criteria, and details of the curation setup.

\section{MulTaBench}
\label{sec:multabench}

MulTaBench is composed of 40 datasets split equally between image-tabular and text-tabular while balancing between regression and classification tasks, all satisfying our curation pipeline established in \S\ref{sec:benchmarking}. The datasets vary in size, ranging from 400 to 114,000 rows and structured feature counts ranging from 1 to 245. While the text-tabular subset is derived exclusively from existing benchmarks, the image-tabular subset is curated and collected from public datasets. Among these, we can find datasets from various domains such as medical and e-commerce. We upload the benchmark to Kaggle\footnote{\url{https://www.kaggle.com/chico89/datasets}}, using a unified API to link between the tables and the images. To facilitate MMTL research, we additionally release 40 candidate datasets meeting the \textit{Joint Signal} criterion, bringing the extended collection to a total of 80 datasets. Appendix~\ref{app:multabench} details the benchmark datasets and their properties.

\begin{figure}[ht]
\centering
\includegraphics[width=0.98\textwidth]{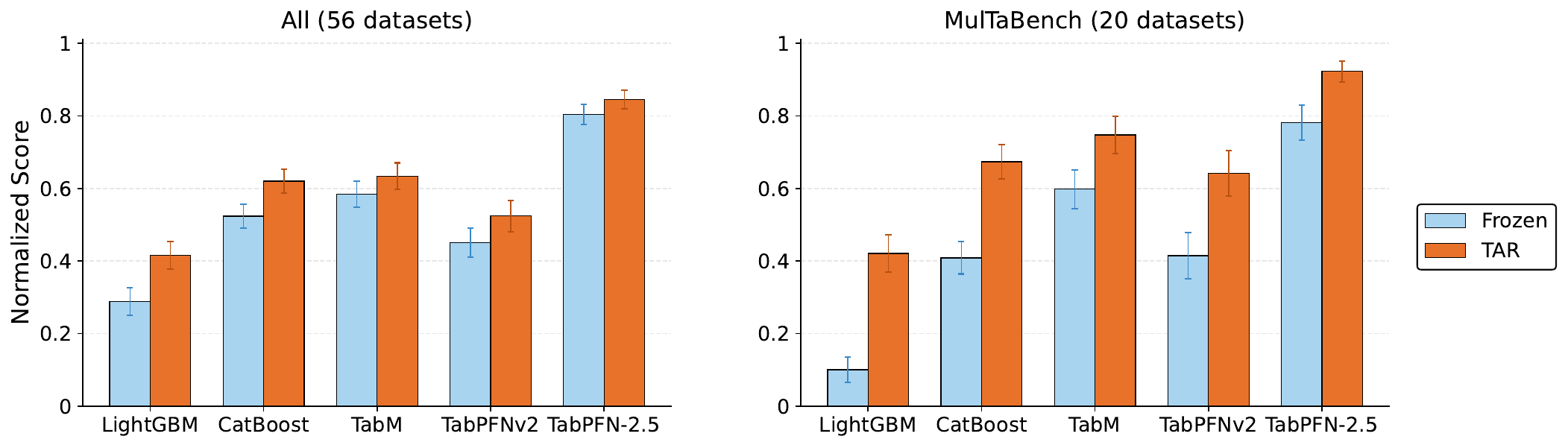}
\caption{Target-Aware Representations Gains over Frozen. Normalized scores for \textit{Joint TAR} and \textit{Joint Frozen} across all text-tabular benchmark datasets (left), and its MulTaBench subset (right).}
\label{fig:text_pool_joint_tar}
\end{figure}

\textbf{Text-Tabular Curation.}\label{par:text_tabular_curation} To evaluate existing text-tabular benchmarks \cite{shi_benchmarking_2021, grinsztajn_vectorizing_2023, kim_carte_2024, mraz_towards_2025}, we aggregate all their 56 unique datasets and subject them to our 4 experimental conditions. To compare classification and regression tasks on a single scale, we normalize AUC and $R^2$ scores to the $[0,1]$ range using min-max scaling and average across datasets, reporting 95\% CIs over all runs. In Figure~\ref{fig:text_pool_joint_tar}, we compare \textit{Joint TAR} and \textit{Joint Frozen} across all datasets, finding that TAR consistently outperforms frozen embeddings for all learners, highlighting the limitations of using fixed representations. Similarly, Appendix~\ref{app:text_curation} shows that the \textit{Joint Signal} condition improves on both \textit{Unimodal} conditions.

With the results in hand, we apply our curation pipeline and find out that approximately 23\% of the datasets fail the \textit{Joint Signal} criterion; of the remaining datasets, 36\% do not pass the \textit{Task-awareness} criterion, leaving 41\% that pass both. From these, we subsample 20 datasets to match the size of the image-tabular subset. Our acceptance rate shows that while our requirements are common enough, they do not constitute the primary focus of standard text-tabular research. Without this distinction, existing benchmarks lack the focus to research target-awareness in MMTL, as shown in Figure~\ref{fig:text_pool_joint_tar}.

\textbf{Image-Tabular Curation.}\label{par:image_tabular_curation} We collect candidate datasets from existing literature \cite{lu_mug_2023, tang_bag_2024, luo_time_2025, kim_multimodalpfn_2025}, identifying a shared pool of 16 unique valid datasets, from which only 5 meet our criteria (31\%), a proportion comparable to the text-tabular subset. We then manually curate additional datasets from Kaggle which pass our pipeline, eventually creating the largest image-tabular benchmark to this date with 20 datasets. In the process, we encountered significant challenges, such as: (1) isolating suitable candidates despite the inconsistent metadata found in public repositories; (2) handling substantial data volumes alongside the fragility of external image links; and (3) determining the appropriate preprocessing for each dataset. Appendix~\ref{app:image_curation} elaborates on the curation process and its difficulties.

\textbf{Trimodal Datasets (Text-Image-Tabular).}\label{par:trimodal_datasets} Within our final image-tabular subset, 9 datasets also contain text fields, leading us to apply the full curation pipeline to the text modality. We evaluate trimodality under a deliberately strict definition: a dataset qualifies as trimodal only if each unstructured modality independently satisfies \textit{Joint Signal} and \textit{Task-awareness}, and adapting both encoders improves upon adapting either one alone. We discover that only 2 datasets satisfy these criteria, with Table~\ref{tab:petfinder} presenting a case study of this analysis for the \textit{PetFinder} dataset. Appendix~\ref{app:trimodal_datasets} discusses whether weaker definitions could apply, noting that the extended collection of 80 released datasets includes 22 potentially trimodal candidates combining text, image, and structured features.

\begin{table}[ht]
\centering
\caption{The PetFinder Analysis. S=Structured, I=Image, T=Text. For all models, performing Joint Modeling and Target-Aware Representations for both modalities maximizes AUC (shown in \%).}

\label{tab:petfinder}
\setlength{\tabcolsep}{4pt}
\small
\begin{tabular}{l cc ccc ccc}
\toprule
 & \multicolumn{2}{c}{\textit{Single modality}} & \multicolumn{3}{c}{\textit{Frozen combinations}} & \multicolumn{3}{c}{\textit{Target-Aware Representations (TAR)}} \\
\cmidrule(lr){2-3}\cmidrule(lr){4-6}\cmidrule(lr){7-9}
\textbf{Model} & I & T & S+I & S+T & S+I+T & S+I\textsubscript{TAR}+T & S+I+T\textsubscript{TAR} & \textbf{S+I\textsubscript{TAR}+T\textsubscript{TAR}} \\
\midrule
LightGBM    & 77.2 & 72.1 & 79.9 & 77.7 & 81.1 & 82.8 & 84.2 & \textbf{85.7} \\
CatBoost    & 78.9 & 73.5 & 81.7 & 79.3 & 83.2 & 83.9 & 85.2 & \textbf{86.4} \\
TabM        & 80.2 & 74.9 & 83.0 & 80.7 & 84.2 & 84.8 & 86.3 & \textbf{87.0} \\
TabPFNv2   & 80.7 & 73.5 & 83.2 & 79.3 & 83.9 & 84.5 & 86.3 & \textbf{87.1} \\
TabPFN-2.5 & 81.1 & 76.0 & 83.7 & 81.0 & 84.9 & 85.3 & 87.3 & \textbf{88.0} \\
\bottomrule
\end{tabular}
\end{table}

\section{Robustness Analysis}\label{sec:results}

While our curation pipeline identifies datasets with high multimodal potential, it is crucial to verify that these properties remain consistent across different modeling choices. We now focus on our second criterion, \textit{Task-awareness}, as it has direct implications for how tabular models process unstructured features. In this section we first ask whether this property generalizes across new tabular learners, embedding models scale, and PCA projection dimensions, and then whether the curated set itself is an artifact of the thresholds and the learners used to build it. Appendix~\ref{app:extended_results} provides extended results and an analysis of the high computational overhead of finetuning (see Appendix~\ref{app:costs}).

\begin{figure*}[h]
    \centering
    \includegraphics[width=0.92\textwidth]{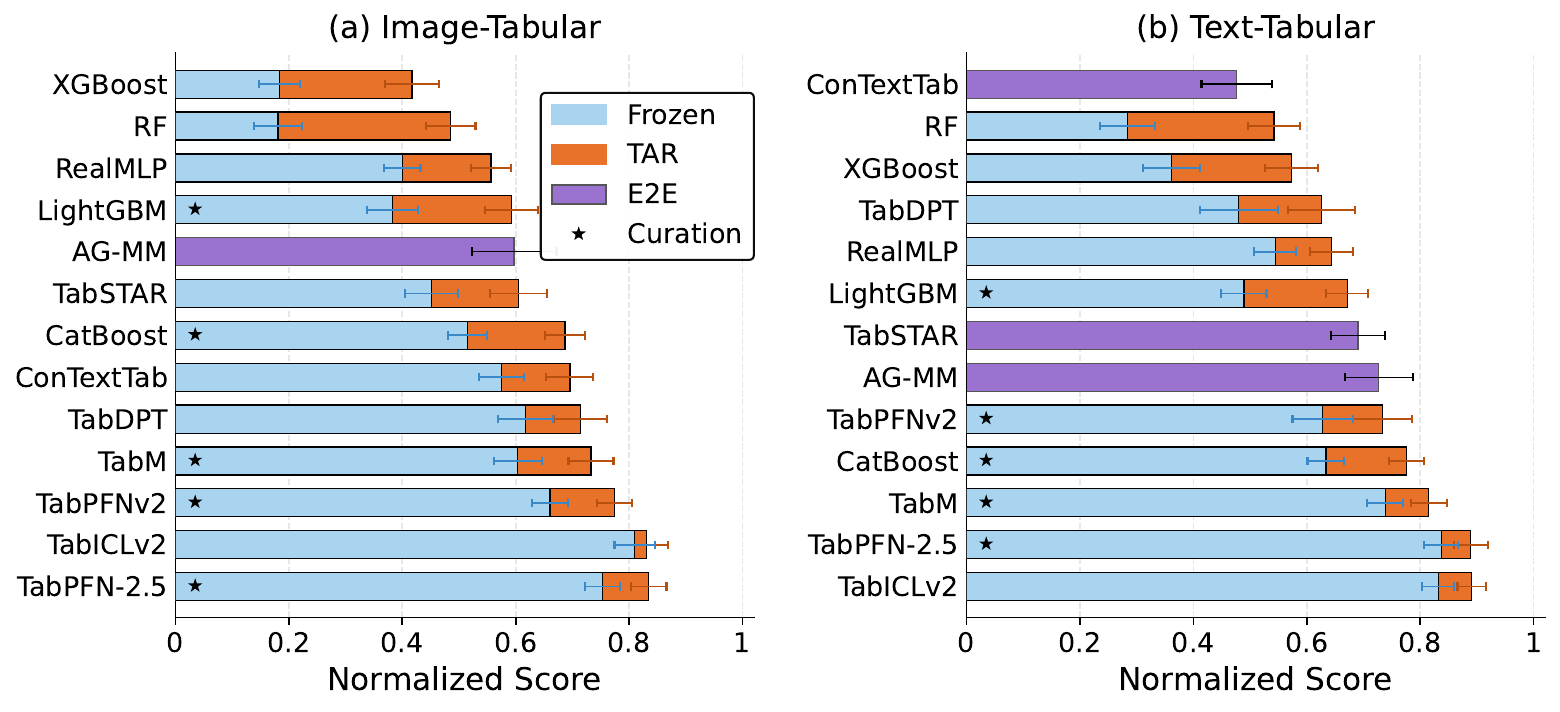}
    \caption{Tabular Learners Performance Analysis. Normalized scores for MulTaBench datasets, with $\pm$ 95\% CI. All learners gain from Target-Aware Representations (TAR).}
    \label{fig:leaderboard}
\end{figure*}

\paragraph{New Tabular Learners.} Since model ranking suffers from selection bias favoring the curation models, our objective is not to establish the SOTA, but to provide a useful tool for the development of future multimodal architectures. Consequently, and to alleviate computation costs, we intentionally omit hyperparameter optimization for both the tabular learners and the TAR preprocessing, underestimating both model performance and the gains from TAR. We supplement the original learners with XGBoost \cite{chen_xgboost_2016}, RandomForest (RF) \cite{breiman_random_2001}, RealMLP \cite{holzmuller_better_2024}, TabDPT \cite{ma_tabdpt_2025}, and TabICLv2 \cite{qu_tabiclv2_2026}. We also include TabSTAR \cite{arazi_tabstar_2025} and ConTextTab \cite{spinaci_contexttab_2025}, which have native text support, and thus are treated as "end-to-end" (E2E) models for the text-tabular subset. Additionally, we evaluate the E2E model AutoGluon-Multimodal (AG-MM) \cite{tang_autogluon-multimodal_2024}, which natively processes texts and images.

Figure~\ref{fig:leaderboard} shows model performance on both MulTaBench subsets. Target-aware embeddings consistently outperform frozen embeddings across all new models and modalities, significantly so on 37 of our 40 datasets (Appendix~\ref{app:results_breakdown}). While this gain is expected for the curation models, its generalization to all the other models provides an indication to the usefulness of our benchmark for MMTL research, confirming the robustness of our curation criteria. We note that GBDTs exhibit the most substantial gains. For the text-tabular subset, ConTextTab is significantly outperformed by AG-MM and TabSTAR, and has the worst performance compared to any TAR variant. This finding is particularly telling, as ConTextTab has set the SOTA for the CARTE Benchmark, emphasizing that MulTaBench targets a fundamentally different text-tabular problem.

\begin{figure*}[h]
\centering
\includegraphics[width=0.90\textwidth]{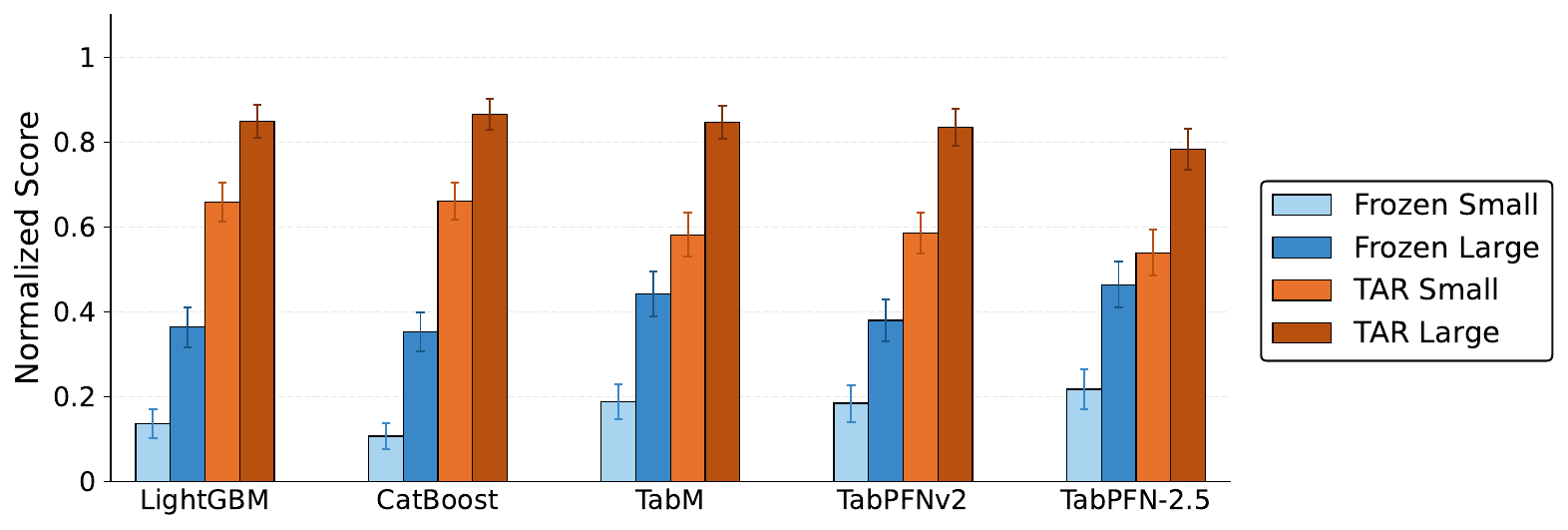}
\caption{Embedding Model Size Analysis. Normalized scores are computed with min-max scaling at the learner level. TAR variants outperform the frozen ones for both model sizes.}
\label{fig:encoder_scale}
\end{figure*}

\paragraph{Embedding Model Scale.} So far, text and image were represented using \textit{e5-v2-small} and \textit{DiNO-v3-small}. Since the dimension of these embeddings is 384, one potential limitation may be that they are too small. We thus repeat the curation experiments using the \textit{Large} variants of the models, which have approximately 10 times more parameters, and a final dimension of 1024. Figure~\ref{fig:encoder_scale} shows that while a larger embedding model improves downstream performance, TAR significantly outperforms frozen embeddings even at the larger scale. In fact, we even observe that the \textit{TAR Small} variant is better than \textit{Frozen Large}; this indicates that increased representational capacity does not guarantee that target-relevant signals are retained in the final representation, and tuning is still required.

\begin{figure*}[h]
\centering
\includegraphics[width=0.80\linewidth]{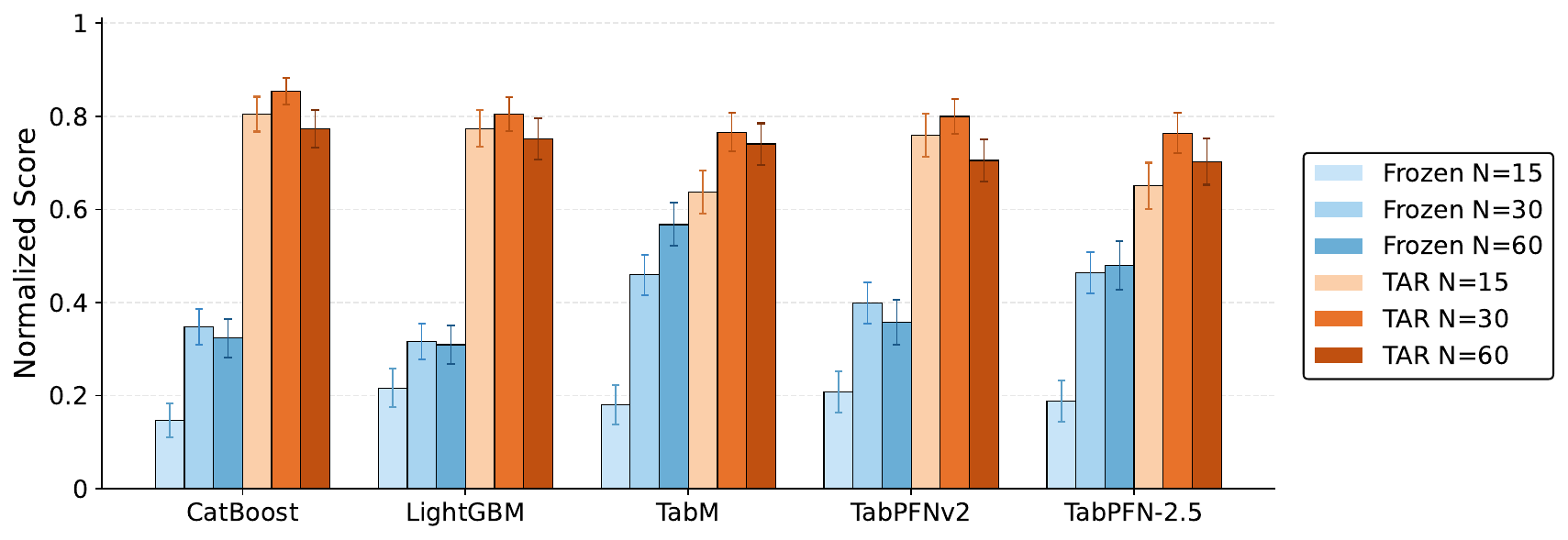}
\caption{Embedding Dimension Analysis. Normalized scores are computed with min-max scaling at the learner level. TAR variants are stronger than Frozen ones for 15, 30, and 60 PCA components.}
\label{fig:pca}
\end{figure*}

\paragraph{Embedding Dimension.} To this point, our analysis has assumed a fixed embedding size of 30 PCA components, following standard practice \cite{grinsztajn_vectorizing_2023, arazi_tabstar_2025}. This dimensionality reduction helps prevent overfitting and ensures computational efficiency by reducing memory requirements. However, this raises the question: is TAR really surfacing information which was missing in the original representations, or is the observed gain an artifact of the compression? We show that representation tuning remains effective across 15 and 60 dimensions (Figure~\ref{fig:pca}), and even when dropping PCA completely in a reduced experiment (Appendix~\ref{app:extended_analysis_no_pca}). This stability proves that our findings are not artifacts of dimensionality, as TAR still improve performance for all of these conditions.

\begin{figure}[h]
\centering
\setlength{\tabcolsep}{1pt}
\renewcommand{\arraystretch}{0.5}
\begin{tabular}{lccc@{\hspace{10pt}}lccc}
 & \small\textbf{Image} & \small\textbf{Frozen} & \small\textbf{Target-Aware} & & \small\textbf{Image} & \small\textbf{Frozen} & \small\textbf{Target-Aware} \\[2pt]
\rotatebox{90}{\small\textbf{CheXpert}} &
\includegraphics[width=0.14\linewidth]{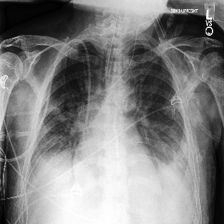} &
\includegraphics[width=0.14\linewidth]{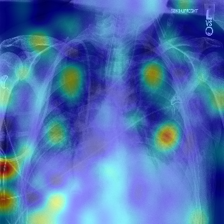} &
\includegraphics[width=0.14\linewidth]{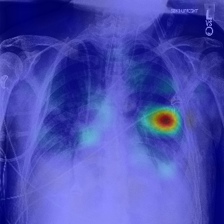} &
\rotatebox{90}{\small\textbf{PetFinder}} &
\includegraphics[width=0.14\linewidth]{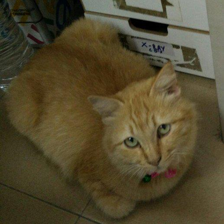} &
\includegraphics[width=0.14\linewidth]{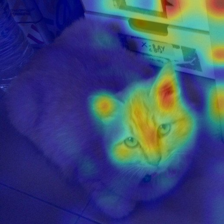} &
\includegraphics[width=0.14\linewidth]{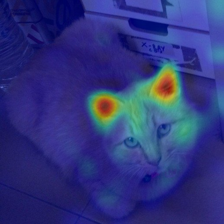} \\[4pt]
\rotatebox{90}{\small\textbf{Glaucoma}} &
\includegraphics[width=0.14\linewidth]{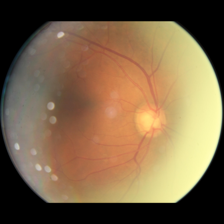} &
\includegraphics[width=0.14\linewidth]{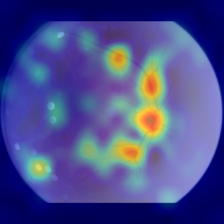} &
\includegraphics[width=0.14\linewidth]{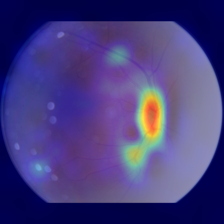} &
\rotatebox{90}{\small\textbf{Celebs}} &
\includegraphics[width=0.14\linewidth]{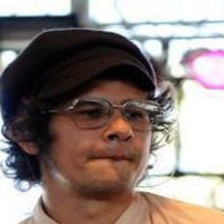} &
\includegraphics[width=0.14\linewidth]{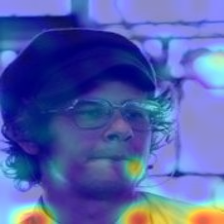} &
\includegraphics[width=0.14\linewidth]{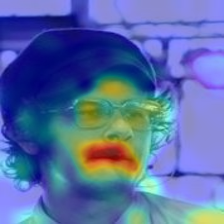} \\
\end{tabular}
\caption{DINO-v3-small Attention Maps. Before (Frozen) and after (Target-Aware) finetuning on the target. The attention shifts from global details to specific regions relevant to the target variable.}
\label{fig:attention_main}
\end{figure}

\paragraph{Qualitative Analysis.} So far, we relied on downstream performance to observe the benefits of TAR on MulTaBench. For image datasets, however, we can extract the last-layer [CLS]-to-patch attention maps from DINO-v3, to see what the representations learn before and after they are adjusted. Figure~\ref{fig:attention_main} illustrates how target-aware adaptation reshapes the encoder's focus across 4 MulTaBench datasets. In \textit{CheXpert} and \textit{Glaucoma}, attention shifts from arbitrary anatomical borders toward the right lower lung and optic disc, respectively. For \textit{PetFinder}, the model suppresses background clutter to focus on the animal, specifically highlighting the ears as key indicators for kitten age. Similarly, focus in \textit{Celebs} moves from peripheral accessories to core facial features. These examples demonstrate that contextualization enables the encoder to surface specific details that are otherwise lost in generic, task-agnostic representations. Appendix~\ref{app:attention_appendix} provides more visual examples.

\paragraph{Curation Pipeline Sensitivity.} Finally, we assess the robustness of our curation methodology, asking whether different hyperparameters and learner panels would have resulted in a different set of accepted datasets. First, we find that the benchmark's composition is robust to the identity of the curation learners. Appendix~\ref{app:curation_committee} shows that when we re-apply the pipeline using any 5-learner subset from our pool of 10, the accepted datasets remain largely unchanged, even when evaluated by a completely disjoint panel. Second, we examine the acceptance criteria. By demanding higher performance gains and broader learner consensus, we observe that dataset acceptance behaves as a spectrum. Appendix~\ref{app:threshold_sensitivity} shows that when we require larger performance improvements or wider agreement among learners, the pipeline gradually isolates the datasets with the strongest need for target-awareness. Overall, while dataset acceptance is robust to the specific models used for curation, stricter thresholding can be used to pick out the datasets where target-awareness matters most.

\section{Towards Multimodal Tabular Foundation Models}\label{sec:priority}

Our analysis of MulTaBench reveals a significant gap between current tabular learners and the demands of MMTL tasks, as existing architectures cannot jointly tune unstructured representations for the target label. In this section, we discuss the potential trajectory of future Multimodal TFMs. Our vision builds upon the framework proposed by \citet{van_breugel_position_2024}. Their position piece identifies TFMs as a research priority and defines 4 core desiderata to guide their development: (D1) handling \textit{mixed-type columns}, such as numbers, categories and dates, (D2) enabling \textit{cross-dataset modeling}, (D3) leveraging \textit{textual context} and metadata, such as column names, and (D4) maintaining \textit{equivariance to column order}. We expand their definition by suggesting (D5) \textit{Target-Aware Multimodal Tabular Learning}; text and image embeddings should be target-aware.

While PFNs have revolutionized structured learning, they are primarily designed for modalities where raw inputs already contain highly compressed signals. Initial efforts attempting to couple PFNs with multimodal encoders \cite{luo_time_2025, kim_multimodalpfn_2025} have struggled to unlock TAR without violating the core ICL premise of avoiding parameter updates. In contrast, joint modeling approaches such as AutoGluon-Multimodal and TabSTAR utilize finetuning to achieve target-awareness, yet this introduces significant practical challenges. Finetuning historically complicates tabular learning by increasing overfitting risks, particularly on small-to-medium datasets, and imposing substantial computational overhead as data, model and embedding scales grow. This burden increases further when using HPO or standard practices like cross-validation and ensembling, as these methods require repeating the expensive finetuning process multiple times to find the best parameters and prevent data leakage across splits.

To summarize, we argue that none of the current architectures are optimal for MMTL, and that the leading paradigms complement each other. MulTaBench enables their development by isolating the datasets that explicitly demand task-specific representations. While proposing a solution is out of this work's scope, we believe that the optimal architecture should take the best of both worlds. An ideal model should bring the contextualization benefits of TAR while preserving the robustness and latency of ICL. We hope that the existence of MulTaBench will enable the research of such models.

\section{Discussion and Conclusion}\label{sec:discussion}

In this work, we introduce MulTaBench, a benchmark of 40 image-tabular and text-tabular datasets designed to explore challenging Multimodal Tabular Learning tasks. We contribute the largest image-tabular benchmark to date, while focusing on tasks that benefit from Joint Modeling and TAR, unlike existing MMTL benchmarks. Our findings show that existing models rely on representations that are often insufficient for the task at hand, making MulTaBench a necessary tool for evaluating the next generation of Multimodal Tabular Foundation Models.

MulTaBench suffers from an important limitation: our curation pipeline entangles the computational problem with the algorithmic solution. As such, it is hard to predict in advance whether a new dataset meets our criteria, and the models used for the curation cannot be fairly evaluated due to selection bias. This entanglement is deliberate: MulTaBench is built to isolate the tasks where frozen representations discard the predictive signal, rather than to rank multimodal tabular models in general. To broaden its scope, we release 40 additional candidate datasets satisfying the \textit{Joint Signal} criterion, yielding a total of 80 datasets to explore MMTL more broadly. Together, they test the property we advocate: a strong architecture should win on MulTaBench without losing ground where frozen encoders already suffice. While we believe future research should aim to address these limitations, our work is a strong step to tackle a problem which was overlooked so far, yielding findings that generalize well to new models. Importantly, the automated nature of our pipeline facilitates the continuous expansion of MulTaBench to include new dataset candidates, the latest tabular learners, or refined selection logic as the field matures. As such, our curation pipeline is a contribution of its own, providing a mechanism to refresh the benchmark with harder candidates as current tasks become saturated by future models.

Our research paves the way to many exciting future directions, such as expanding to a dedicated text-image-tabular benchmark, exploring other modalities such as audio and videos, analyzing different prompting strategies to steer embeddings towards the target, or exploring multimodality for other structured domains, such as relational \cite{robinson_relbench_2024, hayler_relarena_2026} or time-series \cite{shchur_fevbench_2025}. Mainly, MulTaBench supports the development of Multimodal TFMs. In our opinion, there are two big challenges to solve: architecture and training data. For architectures, in \S\ref{sec:priority}, we suggest that future models should ideally take the best out of ICL and finetuning; for instance, coupling TFMs with LLMs and VLMs is a compelling path. For training data, since real data corpora for MMTL are rare \cite{eggert_tablib_2023}, expanding the synthetic numerical priors used for training TFMs \cite{hollmann_accurate_2025, zhang_mitra_2025, qu_tabiclv2_2026} to include text and image features is an exciting direction \cite{luo_can_2025, brahmavar_task_2026}. We hope that our work will contribute to the research of Multimodal Tabular Learning, and we are excited towards a future where this crucial problem sees the progress it deserves.

\begin{ack}
AA, ES, SG, MV, and RR are supported by an Israel Ministry of Science and Technology (MOST) grant on multi-modal AI. ES is supported by a Google PhD Fellowship. GV acknowledges support from ANR via grant TaFoMo (ANR-25-CE23-1822). This work is partly supported by Hi! PARIS and ANR/France 2030 program (ANR-23-IACL-0005). FH acknowledges the financial support of the Hector Foundation.
LP acknowledges funding by the Deutsche Forschungsgemeinschaft (DFG, German Research Foundation) under SFB 1597 (SmallData), grant number 499552394.
Funded by the European Union. Views and opinions expressed are however those of the author(s) only and do not necessarily reflect those of the European Union or the European Commission. Neither the European Union nor the European Commission can be held responsible for them. This work was supported by the European Union’s Horizon Europe research and innovation programme under grant agreement No 101214398 (ELLIOT).
We acknowledge the EuroHPC Joint Undertaking for awarding this project access to the EuroHPC supercomputer LUMI, hosted by CSC (Finland) and the LUMI consortium through a EuroHPC AI Factory Large Scale Access call (project EHPC-AIF-2025LS17-014). We acknowledge EuroHPC JU for awarding the project ID EHPC-AIF-2025LS14-031 access to MareNostrum 5 ACC hosted by the Barcelona Supercomputing Center (BSC), Spain.
\begin{center}
\begin{minipage}{0.3\textwidth}
\centering
\includegraphics[width=\textwidth]{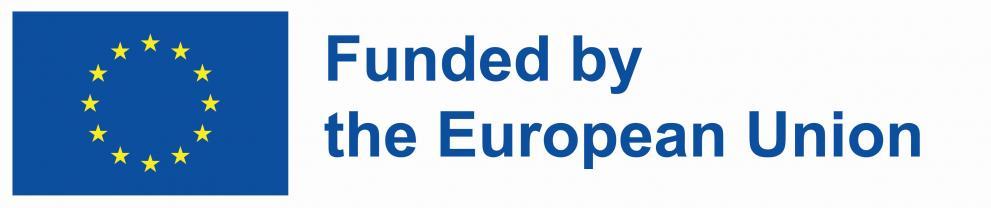}
\end{minipage}
\begin{minipage}{0.3\textwidth}
\centering
\includegraphics[width=\textwidth]{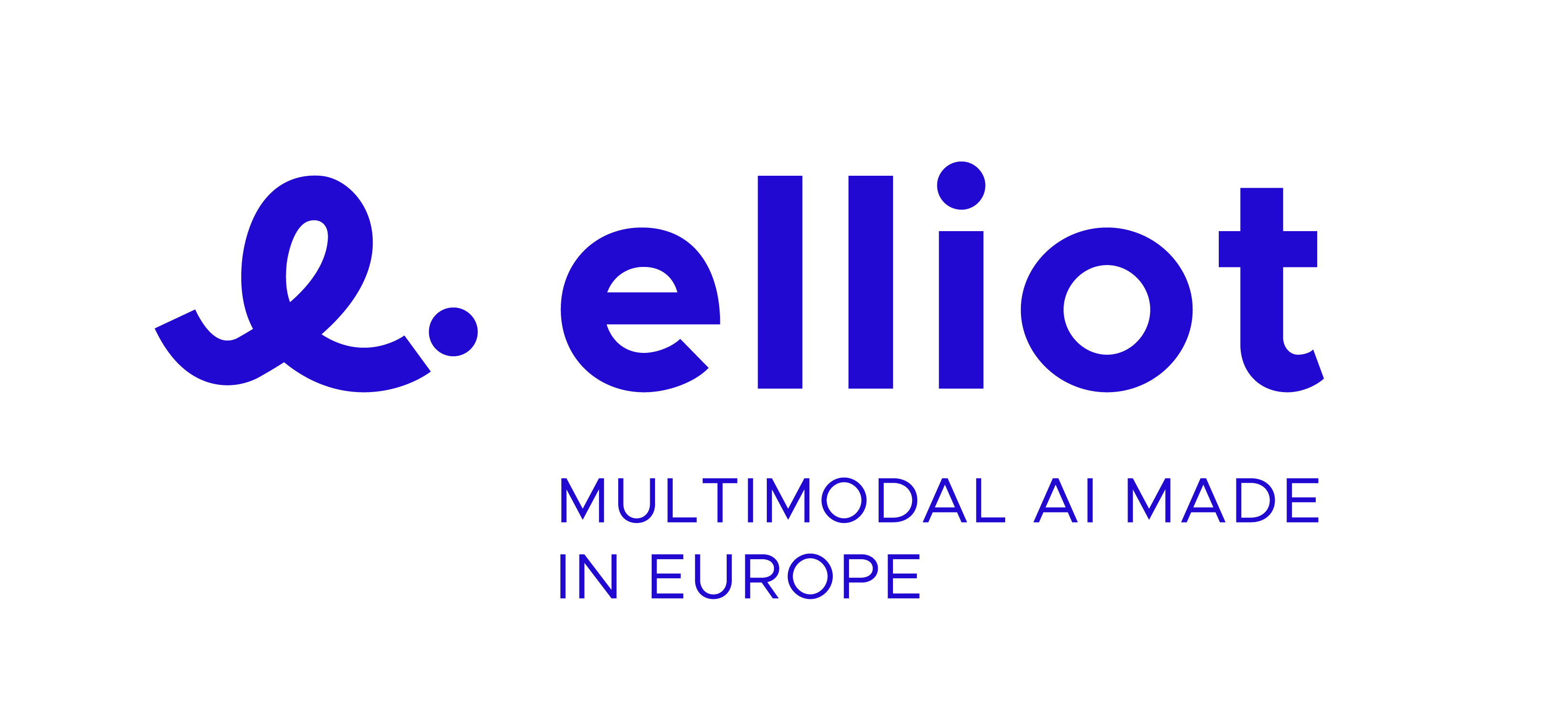}
\end{minipage}
\begin{minipage}{0.3\textwidth}
\centering
\includegraphics[width=\textwidth]{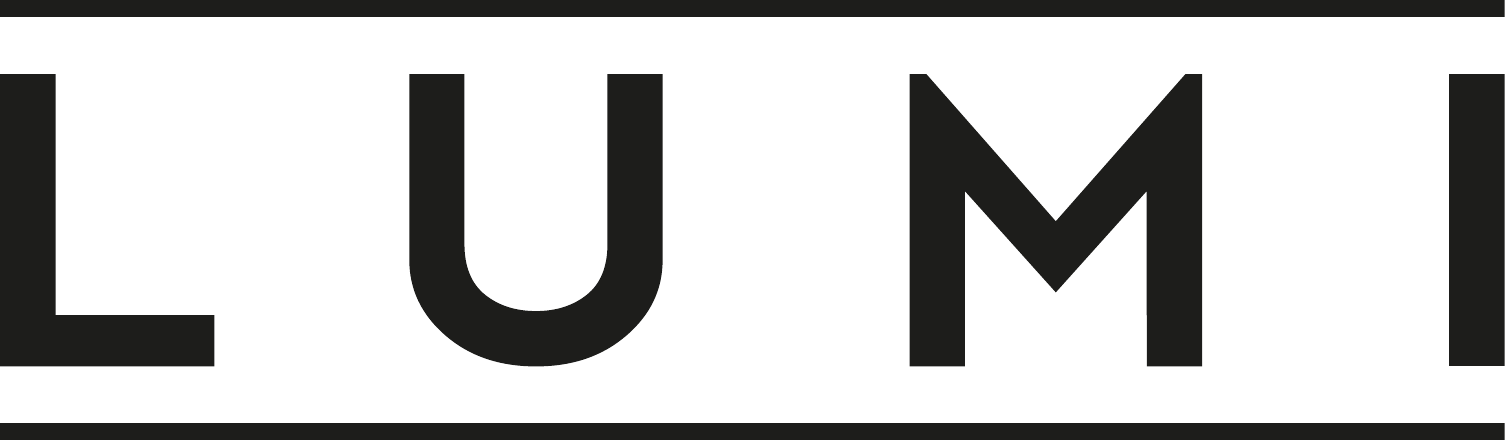}
\end{minipage}
\end{center}

\textbf{Competing interests.} AA, ES, and RR are the developers of TabSTAR. DH is one of the authors of RealMLP. LP and FH are authors of TabPFNv2 and TabPFN-v2.5. DH and GV are authors of TabICLv2. LP and DH are maintainers of TabArena. LP is a maintainer of AutoGluon. GV is a core contributor to skrub and a co-founder of scikit-learn. AA, LP, and FH are affiliated with Prior Labs, a company developing the TabPFN model family. GV is affiliated with Probabl, a company developing software for industrial-grade data science. The authors declare no other competing interests.

\end{ack}

\bibliographystyle{plainnat}
\bibliography{overrides, neurips2026}

\clearpage
\appendix

\section{Curation Pipeline}
\label{app:curation}

\subsection{Target-Aware Representations}
\label{app:curation_tar_training}

Target-Aware Representations are produced by finetuning the top 3 transformer layers of the encoder using LoRA \cite{hu_lora_2021}, with a single linear head mapping the encoder output (384-dim) to the number of output classes. Finetuning is performed as a preprocessing step, independently of the structured features and the downstream tabular learner. The encoder is adapted on the training split only, using a stratified 90/10 train/validation split to select the best checkpoint. Importantly, there is no data leakage, as the test set is never used for this step, just like any other preprocessing.

\paragraph{Hyperparameters.} Both DINO-v3-small\footnote{\url{https://huggingface.co/facebook/dinov3-vits16-pretrain-lvd1689m}} and e5-small-v2\footnote{\url{https://huggingface.co/intfloat/e5-small-v2}} share the same LoRA configuration: $r=16$, $\alpha=32$, dropout $0.1$. Training uses AdamW with learning rate $10^{-4}$ for e5 and $0.001$ for DINO, with a batch size of 256, and weight decay $0.01$. For DINO, we train to up to 100 epochs. As many datasets have multiple text features, we reduce this number for e5 to 50. We apply early stopping after 3 epochs of no improvement on the validation loss. All hyperparameters are fixed across datasets; no per-dataset tuning is performed. Reported gains are therefore conservative lower bounds on what task-specific adaptation could achieve.

\paragraph{Regression.} For regression targets, the continuous label is discretized into 20 equal-frequency bins and the adaptation objective is cross-entropy over these bins. We find this technique to be more stable than direct regression finetuning, as it is much less sensitive to outliers. However, it's plausible that this decision could be optimized much further.

\paragraph{Text.} While MulTaBench image datasets have a single image feature, text-tabular datasets often have more than one text field, which we defined as string features that have at least 100 distinct values. For efficiency, a single e5 model is finetuned jointly across all text columns: each row-col pair generates one training example in the format ``$col\_name: col\_val$", paired with the row's target label. This allows the model to learn a shared representation across all text features simultaneously. This decision might harm representations, especially as feature size grows, but finetuning a dedicated embedding model for each feature would have been computationally infeasible.

\subsection{Curation Experimental Setup}

Each candidate dataset is evaluated by 5 tabular learners: LightGBM, CatBoost, TabM, TabPFNv2, and TabPFN-2.5, over five random seeds under the 4 conditions defined in \S\ref{sec:benchmark_design}. Training is capped at 10,000 examples per fold, and the metric is AUC for classification and $R^2$ for regression tasks.

We run models using default configurations. For LightGBM, we use its default implementation\footnote{\url{https://pypi.org/project/lightgbm/}}. For CatBoost, we follow previous work \cite{gorishniy_revisiting_2021, arazi_tabstar_2025} and set $early\_stopping\_rounds = 50, od\_pval = 0.001, iterations = 2000$. For TabM, we use its \textit{pytabkit}\footnote{\url{https://github.com/dholzmueller/pytabkit}} implementation with default parameters. For TabPFNv2 and TabPFN-2.5, we use their default implementation.\footnote{\url{https://github.com/PriorLabs/TabPFN}}

\subsection{Formal Acceptance Criteria}
\label{app:curation_formal}

Let $\mathcal{D}$ be a candidate dataset and $\mathcal{M}$ be a pool of 5 curation tabular learners. For a given learner $m \in \mathcal{M}$, let $S_m(\text{Condition})$ denote its average predictive performance (AUC or $R^2$) under a given condition.

\paragraph{Joint Signal.} We define the \textit{Joint} gain as the improvement of the joint model over the strongest unimodal baseline:

$$ \Delta_{\text{Joint}}(m) = S_m(\text{Joint Frozen}) - \max\big(S_m(\text{UnimodalStructured}),\; S_m(\text{UnimodalUnstructured})\big) $$

\paragraph{Task-awareness.} We define the \textit{Awareness} gain as the improvement of Joint TAR over Joint Frozen:

$$ \Delta_{\text{Awareness}}(m) = S_m(\text{Joint TAR}) - S_m(\text{Joint Frozen}) $$

\paragraph{Selection rule.} To ensure that the observed improvements are robust and meet a minimum significance margin, we introduce a threshold parameter $\delta > 0$ and a consensus fraction $\rho \in (0.5, 1]$. A dataset $\mathcal{D}$ is accepted if and only if both gains reach $\delta$ for a majority of the learners:

$$ \text{Accept}(\mathcal{D}) \iff \left|\left\{ m \in \mathcal{M} \;:\; \Delta_{\text{Joint}}(m) \geq \delta \;\land\; \Delta_{\text{Awareness}}(m) \geq \delta \right\}\right| \geq \rho \cdot |\mathcal{M}| $$

The two conditions are evaluated jointly per learner: a model counts toward the threshold only if both gains reach it. Scores are averaged across folds and rounded to three decimals, so every gain is a multiple of $0.001$ and $\delta$ is the smallest improvement the rule can express. In our case, we set $\delta=0.001$ and $\rho=3/5$. We note that since we use a binary decision threshold, some datasets can marginally cross it while others can generate consensus. Demanding stricter thresholds could enhance the robustness of the selected datasets. Appendix~\ref{app:threshold_sensitivity} quantifies how much the selected set depends on both thresholds.

\section{MulTaBench Datasets}
\label{app:multabench}

In this section we present MulTaBench datasets. Table~\ref{tab:multabench_datasets} provides their high-level statistics, including the number of rows and feature type breakdown. The rest of the section provides a concise per-dataset high-level description; their exact preprocessing logic can be found in our released code.

\subsection{Image-Tabular Dataset Descriptions}

\begin{table*}[!t]
\centering
\caption{All 40 MulTaBench Datasets Properties. \textit{Task}: Classification (CLS) or Regression (REG). \textit{Classes}: number of target classes (for CLS). \textit{N}: total examples. \textit{Struct.}: numerical + categorical features. \textit{Text}: free-text features. \textit{Img.}: image features.}
\label{tab:multabench_datasets}
\footnotesize
\begin{tabular}{llrrrrr}
\toprule
Dataset & Task & Classes & $N$ & Struct. & Text & Img. \\
\midrule
\multicolumn{7}{l}{\textit{Image-Tabular (20 datasets)}} \\
\midrule
CBIS-DDSM             & CLS &   4 &   1,696 &   8 &  0 & 1 \\
Celeb Attractiveness  & CLS &   2 &  99,999 &  39 &  0 & 1 \\
CheXpert              & CLS &   3 &  46,437 &  17 &  0 & 1 \\
CS:GO Skins           & CLS &  10 &     956 &   3 &  1 & 1 \\
Flower Bouquets       & CLS &   5 &     600 &   3 &  1 & 1 \\
Glaucoma SMDG         & CLS &   3 &  12,449 &   8 &  0 & 1 \\
Hateful Meme          & CLS &   2 &   9,000 &  20 &  0 & 1 \\
HubMAP HPA            & CLS &  10 &  12,581 &   5 &  1 & 1 \\
Justin Instagram      & CLS &   5 &  10,319 &   6 &  0 & 1 \\
Mammography CMMD      & CLS &   2 &   5,202 &   4 &  0 & 1 \\
PetFinder             & CLS &   8 &  14,652 &  17 &  4 & 1 \\
Zooscan Plankton      & CLS &  10 & 100,000 &  28 &  0 & 1 \\
Amazon Bestseller     & REG &  -- &   3,488 &   4 &  0 & 1 \\
Amazon Packages       & REG &  -- &  46,398 &   1 &  1 & 1 \\
H\&M Fashion          & REG &  -- & 104,072 &   9 &  4 & 1 \\
Khaadi Clothes        & REG &  -- &     400 &   2 &  1 & 1 \\
Letterboxd Movies     & REG &  -- &  12,564 &  23 &  3 & 1 \\
Mango Mass            & REG &  -- &     546 &   2 &  0 & 1 \\
MkPhoto Bots          & REG &  -- &  13,748 &   8 &  0 & 1 \\
Painting Price        & REG &  -- &  12,369 & 245 &  2 & 1 \\
\midrule
\multicolumn{7}{l}{\textit{Text-Tabular (20 datasets)}} \\
\midrule
Data Scientist Salary & CLS &   6 &  15,841 &   1 &  5 & 0 \\
Fake Job Postings     & CLS &   2 &  12,725 &   2 &  3 & 0 \\
Jigsaw Toxicity       & CLS &   2 & 100,000 &  30 &  1 & 0 \\
Kickstarter           & CLS &   2 &  86,502 &   6 &  3 & 0 \\
Michelin Guide        & CLS &   5 &  18,843 &   5 &  6 & 0 \\
Product Sentiment     & CLS &   4 &   5,091 &   1 &  1 & 0 \\
Spotify Genres        & CLS & 114 & 114,000 &  15 &  3 & 0 \\
US Accidents          & CLS &   4 & 100,001 &  38 &  6 & 0 \\
Wine Review           & CLS &  30 &  84,123 &   3 &  2 & 0 \\
Women's Clothing      & CLS &   5 &  18,788 &   8 &  2 & 0 \\
Baby Products         & REG &  -- &   5,085 &   8 &  4 & 0 \\
Book Price            & REG &  -- &   4,989 &   3 &  5 & 0 \\
Book Readability      & REG &  -- &   4,724 &  24 &  6 & 0 \\
Mercari Marketplace   & REG &  -- & 100,000 &   3 &  6 & 0 \\
Montgomery Salaries   & REG &  -- &   9,228 &   8 &  3 & 0 \\
Rotten Tomatoes       & REG &  -- &   7,158 &   2 & 13 & 0 \\
SciMagojr Impact      & REG &  -- &  31,136 &  12 & 10 & 0 \\
Vancouver Salaries    & REG &  -- &  44,574 &   3 &  2 & 0 \\
Video Games Sales     & REG &  -- &  16,598 &   3 &  2 & 0 \\
Zomato Restaurants    & REG &  -- &  41,665 &   8 &  7 & 0 \\
\bottomrule
\end{tabular}
\end{table*}

\paragraph{\href{https://www.kaggle.com/datasets/awsaf49/cbis-ddsm-breast-cancer-image-dataset}{CBIS-DDSM.}} Cropped mammography mass regions from the Curated Breast Imaging Subset of DDSM.
The 4-class target is BI-RADS breast density.
Structured features describe lesion morphology, such as laterality, imaging view, mass shape and margins, BI-RADS assessment score, pathology, and subtlety rating.

\paragraph{\href{https://www.kaggle.com/datasets/jessicali9530/celeba-dataset}{Celeb Attractiveness.}} Celebrity face images from the CelebA dataset, subsampled from the full release.
The binary target is a crowd-annotated attractiveness label.
Each row pairs the face image with binary facial-attribute flags such as \textit{Smiling}, \textit{Wearing\_Lipstick}, or \textit{Bald}, making the image a complement to an already rich structured signal.

\paragraph{\href{https://www.kaggle.com/datasets/ashery/chexpert}{CheXpert.}} Chest X-ray images from the Stanford CheXpert dataset, covering frontal and lateral views.
The 3-class target is the Cardiomegaly label (positive, negative, or uncertain).
Structured features include patient sex, age, and co-occurring pathology labels, many of them sparsely observed, reflecting the natural label uncertainty of radiology reports.

\paragraph{\href{https://figshare.com/ndownloader/files/38077458}{CS:GO Skins.}} Weapon skin images and marketplace metadata from Counter-Strike: Global Offensive.
The 10-class target discretizes market price into decile bins.
Structured features include the skin rarity tier, weapon category, and availability, with a free-text skin name describing the visual design.

\paragraph{\href{https://www.kaggle.com/datasets/olgabelitskaya/flower-color-images}{Flower Bouquets.}} Flower bouquet photographs paired with sales metadata from a Russian online florist.
The 5-class target is a customer satisfaction rating (1--5).
Features include a free-text bouquet description, the average comment-based rating, and the price.

\paragraph{\href{https://www.kaggle.com/datasets/deathtrooper/multichannel-glaucoma-benchmark-dataset}{Glaucoma SMDG.}} Retinal fundus photographs from the SMDG multi-source glaucoma benchmark.
The 3-class target encodes glaucoma diagnosis (positive, negative, or uncertain).
Clinical metadata such as patient age, sex, laterality, and intraocular pressure are available but heavily sparse, reflecting real-world incompleteness in ophthalmic records.

\paragraph{\href{https://www.kaggle.com/datasets/parthplc/facebook-hateful-meme-dataset}{Hateful Meme.}} Multimodal memes from the Facebook Hateful Memes Challenge.
The binary target labels each meme as hateful or not.
To make it a tabular task, the text field is pre-embedded into continuous variables, so the structured columns should be treated as numeric features rather than raw text, with the meme image providing complementary visual context.

\paragraph{\href{https://www.kaggle.com/datasets/miquel0/hubmaphba-tiled-dataset-512x512}{HubMAP HPA.}} Histology tissue tiles from the HuBMAP-HPA organ segmentation competition.
The 10-class target discretizes donor age into decile bins, asking whether tissue morphology encodes biological age.
Structured features include the organ type, donor sex, the tile index, and the source image dimensions.

\paragraph{\href{https://www.kaggle.com/datasets/aldiandyainf/which-justin-posted-that}{Justin Instagram.}} Instagram posts from five celebrities named Justin (Bieber, Trudeau, Timberlake, Long, Hartley).
The 5-class target identifies which Justin authored each post.
Structured features include post-level metadata: counts of hashtags, characters, words, emojis, and mentions, plus a binary video indicator.

\paragraph{\href{https://www.kaggle.com/datasets/nguynththanhho/cmmd-mammography}{Mammography CMMD.}} Mammography images from the Chinese Mammography Database, cropped to the lesion region.
The binary target distinguishes malignant from benign findings.
Structured features include patient age, laterality, abnormality type (mass, calcification, or both), and the cropping method used.

\paragraph{\href{https://www.kaggle.com/datasets/c/petfinder-adoption-prediction}{PetFinder.}} Pet adoption listings from the Malaysian PetFinder platform.
The 8-class target discretizes listed pet age into octile bins, testing whether visual appearance and listing text jointly predict developmental stage.
Features include the species, breed, color, health status, adoption fee, and state location, alongside a free-text listing description and the pet's photograph.

\paragraph{\href{https://www.kaggle.com/datasets/raghavdharwal/pelgass-bay-of-biscay-zooscan-zooplankton-dataset}{Zooscan Plankton.}} Zooplankton specimens from the PELGAS Bay of Biscay survey, scanned with a ZooScan optical system.
The 10-class target classifies copepod taxa such as Calanoida, Oithonidae, Calanidae, and Temoridae.
Structured features include morphometric descriptors computed from the scan, such as circularity, skewness, and fractal dimension, alongside sampling metadata such as geographic coordinates, depth, collection date, and mesh size.

\paragraph{\href{https://www.kaggle.com/datasets/amankumar20d/amazon-best-seller-all-departments-us}{Amazon Bestseller.}} Product listings from Amazon's bestseller rankings across all departments.
The target is the log-transformed product price.
Structured features include the number of ratings, the bestseller rank within department, the star rating, and the list page, while the thumbnail image gives visual cues about item type and packaging.

\paragraph{\href{https://www.kaggle.com/datasets/dhruvildave/amazon-bin-image-dataset}{Amazon Packages.}} Warehouse bin images from Amazon's robotic fulfillment centers.
The target is the total weight of the bin's contents in pounds.
The only structured feature is the expected item count, with a free-text product description naming the item in each bin.

\paragraph{\href{https://www.kaggle.com/datasets/odins0n/handm-dataset-128x128}{H\&M Fashion.}} Clothing article metadata and thumbnail images from H\&M's product catalog.
The target is the average age of purchasing customers, capturing whether visual style and descriptive text encode demographic appeal.
Structured attributes are the product type, color group, graphical appearance, garment group, and department, with the product name and a free-text detail description as text features, making this a trimodal dataset.

\paragraph{\href{https://www.kaggle.com/datasets/usman8/khaadis-clothes-data-with-images}{Khaadi Clothes.}} Apparel listings from the Pakistani fashion brand Khaadi.
The target is the retail price in Pakistani rupees.
Structured features include the color and product category, with a free-text description specifying fabric type and construction.

\paragraph{\href{https://www.kaggle.com/datasets/gsimonx37/letterboxd}{Letterboxd Movies.}} Film metadata and poster images from the Letterboxd movie-tracking platform, covering releases between 2021 and 2024.
The target is the average community rating.
Features include binary genre flags, the release year, and the runtime, with the tagline and theme descriptions as text alongside the official poster image.

\paragraph{\href{https://www.kaggle.com/datasets/saurabhshahane/mango-varieties-classification}{Mango Mass.}} Mango fruit photographs from a variety classification study.
The target is the fruit mass in kilograms.
The only structured features are the color group and the quality grade, making the image the dominant signal for weight prediction.

\paragraph{\href{https://www.kaggle.com/datasets/guardeec/mkphoto2023}{MkPhoto Bots.}} Social media photographs collected for image authenticity research.
The target is a continuous trust score, the estimated probability that the post is genuine.
Structured features include binary flags for GAN generation and deepfake manipulation, presence of a person, face count, a recognized-celebrity list, upload speed, and a noise-quality score.

\paragraph{\href{https://www.kaggle.com/datasets/denozavrus/paintings-price-prediction}{Painting Price.}} Painting images and metadata from an online art marketplace.
The target is the sale price.
Structured features include the physical dimensions, the material, and binary style tags such as \textit{abstract} or \textit{impressionism}, with a high-cardinality free-text styles column adding further stylistic signal.

\subsection{Text-Tabular Dataset Descriptions}

\paragraph{\href{https://www.openml.org/search?type=data&id=46664}{Data Scientist Salary.}} Indian data science job postings.
The 6-class target is the salary band in lakh rupees per annum.
Text features include the experience range, job description, designation, required key skills, and city, with a noisy job-type field as a weak structured signal.

\paragraph{\href{https://www.openml.org/search?type=data&id=46655}{Fake Job Postings.}} Job listings annotated for authenticity.
The binary target flags fraudulent listings.
Text features include the job title and full description, while structured features capture the required experience level, required education, and salary range, testing whether deceptive intent surfaces in free text beyond coarse metadata.

\paragraph{\href{https://www.openml.org/search?type=data&id=46654}{Jigsaw Toxicity.}} Online comments from the Civil Comments platform, collected for Jigsaw's toxicity detection task.
The binary target labels each comment as toxic.
Alongside the comment text, structured features include identity-mention fraction scores, community-reaction counts, and the comment creation timestamp.

\paragraph{\href{https://www.openml.org/search?type=data&id=46668}{Kickstarter.}} Crowdfunding campaigns from Kickstarter.
The binary target indicates whether the funding goal was reached.
Text features include the project name, description, and keyword slug, while structured features cover the funding goal, country, currency, and the campaign deadline and creation timestamps.

\paragraph{\href{https://www.kaggle.com/datasets/ngshiheng/michelin-guide-restaurants-2021}{Michelin Guide.}} Restaurant listings from the 2021 Michelin Guide.
The 5-class target is the award level, from Selected Restaurants and Bib Gourmand up to three stars.
Text features include the restaurant name, address, location, cuisine type, facilities, and the Michelin editorial description, while structured features cover the geo-coordinates, price tier, and a Green Star sustainability flag.

\paragraph{\href{https://www.openml.org/search?type=data&id=46651}{Product Sentiment.}} Tweets about Apple, Google, and Twitter products posted during SXSW 2011.
The 4-class target is sentiment: Positive, Negative, No Sentiment, or Cannot Say.
The only text feature is the tweet content, with an integer-encoded product-type column identifying the product being discussed.

\paragraph{\href{https://www.kaggle.com/datasets/maharshipandya/-spotify-tracks-dataset}{Spotify Genres.}} Spotify track metadata sampled evenly across genres.
The 114-class target is the track genre.
Text features include the artist, album, and track name, while structured features cover the Spotify audio descriptors such as danceability, energy, loudness, and valence.

\paragraph{\href{https://www.kaggle.com/datasets/sobhanmoosavi/us-accidents}{US Accidents.}} Traffic accident records from the contiguous United States.
The 4-class target is accident severity on a 1--4 scale.
Text features include a free-text incident description and several location fields, while structured features cover the coordinates, state and timezone, weather measurements and conditions, timestamps, and binary road-feature flags.

\paragraph{\href{https://www.openml.org/search?type=data&id=46653}{Wine Review.}} Professional wine tasting notes from Wine Enthusiast magazine.
The 30-class target is the grape variety.
Text features include the tasting note and the province of origin, with the numeric rating, price, and country as structured features, making grape identification from flavor language a natural text-tabular benchmark.

\paragraph{\href{https://www.openml.org/search?type=data&id=46659}{Women's Clothing.}} Customer reviews of women's clothing from an anonymous US e-commerce retailer.
The 5-class target is the star rating (1--5).
Text features include the review title and body, while structured features cover the customer age, product and department metadata, a recommendation indicator, and the positive feedback count.

\paragraph{\href{http://pages.cs.wisc.edu/~anhai/data/784_data/baby_products/}{Baby Products.}} Nursery and baby product listings from a US retail catalog.
The target is the retail price.
Text features include the product title, brand name, and sparsely populated color, fabric, and material fields, while structured features cover a discount flag, the product category, and physical dimensions.

\paragraph{\href{https://www.openml.org/search?type=data&id=46663}{Book Price.}} Books listed on an online marketplace.
The target is the log-transformed price in USD.
Text features include the title, author, edition details, synopsis, genre tag, and broad book category, with the average star rating and the number of ratings as structured features.

\paragraph{\href{https://www.kaggle.com/datasets/verracodeguacas/clear-corpus}{Book Readability.}} Text excerpts from the CLEAR Corpus of children's and educational literature.
The target is the New Dale--Chall Readability Formula score, a standard measure of text difficulty.
The excerpt itself is the key text feature, accompanied by linguistic and bibliographic attributes such as the publication year, sentence and paragraph counts, classical readability metrics, and MPAA content rating.

\paragraph{\href{https://www.openml.org/search?type=data&id=46660}{Mercari Marketplace.}} Secondhand item listings from the Mercari mobile marketplace.
The target is the log-transformed sale price.
Text features include the item name, the free-text description, and a hierarchical category label, while structured features cover the item condition, the brand name, and a shipping-included flag.

\paragraph{\href{https://www.openml.org/search?type=data&id=42125}{Montgomery Salaries.}} Annual salary records of Montgomery County (Maryland, USA) government employees.
The target is the current annual salary.
Text features include the employee name, division, and job title, while structured features cover gender, prior-year gross and overtime pay, department, assignment type, and hire date.

\paragraph{\href{http://pages.cs.wisc.edu/~anhai/data/784_data/movies1/}{Rotten Tomatoes.}} Movie metadata aggregated from IMDb and Rotten Tomatoes.
The target is an audience and critic composite rating.
Text features include the movie name, director, screenwriter, cast list, language, country, filming locations, genre tags, and plot description, while structured features cover the release year, runtime, and rating and review counts.

\paragraph{\href{https://www.scimagojr.com/journalrank.php}{SciMagojr Impact.}} Academic journal and book series metadata from the SCImago Journal \& Country Rank database.
The target is the journal's H-index.
Text features include the title, publisher, coverage period, and subject categories, while structured features cover the SJR impact score, quartile ranking, document and citation counts, Overton policy citation index, and SDG alignment score.

\paragraph{\href{https://opendata.vancouver.ca}{Vancouver Salaries.}} Annual salary disclosures for City of Vancouver public employees.
The target is the annual remuneration.
Text features include the job title and department name, with the fiscal year, employee name, and declared expenses as structured features.

\paragraph{\href{https://www.kaggle.com/datasets/gregorut/videogamesales}{Video Games Sales.}} Video game sales records from VGChartz.
The target is the global sales in millions of units.
Text features include the game title and publisher name, with the platform, release year, and genre as structured features.

\paragraph{\href{https://www.kaggle.com/datasets/himanshupoddar/zomato-bangalore-restaurants}{Zomato Restaurants.}} Restaurant listings from the Zomato platform covering Bangalore, India.
The target is the aggregate user rating.
Text features include the restaurant name, address, cuisine types, highlighted dishes, raw user review text, and menu item lists, while structured features cover the online ordering and reservation availability, total votes, neighborhood, restaurant type, and approximate cost for two.

\section{Text-Tabular Curation}
\label{app:text_curation}

We evaluate existing text-tabular benchmarks by drawing candidates from 4 sources: the Multimodal AutoML Benchmark \cite{shi_benchmarking_2021}, \citet{grinsztajn_vectorizing_2023}, CARTE \cite{kim_carte_2024}, and TextTabBench \cite{mraz_towards_2025}, yielding 56 unique candidates after deduplication and exclusion of datasets which were unavailable due to improper hosting. Each dataset is evaluated by 5 tabular learners over 5 folds under the 4 conditions defined in \S\ref{sec:benchmark_design}.

\subsection{Existing Benchmarks}

The 4 source benchmarks share a substantial number of datasets, either by directly using the exact same source or by using similar-enough datasets. We adopt the deduplication performed by \citet{arazi_tabstar_2025}, and extend it to include TextTabBench \cite{mraz_towards_2025}, yielding a pool of 56 unique datasets. Table~\ref{tab:text_deduplication} shows datasets which are shared across more than one existing text-tabular benchmark.

\begin{table}[ht]
\centering
\caption{Duplicate datasets across benchmarks. \checkmark\ indicates presence.}
\label{tab:text_deduplication}
\small
\begin{tabular}{lcccc}
\toprule
\textbf{Dataset} & \textbf{AutoML Multimodal} & \textbf{Grinsztajn et al} & \textbf{CARTE} & \textbf{TextTabBench} \\
\midrule
Wine Review         & $\checkmark$ & $\checkmark$ & $\checkmark$ & $\checkmark$ \\
BikeWale Bikes      &              & $\checkmark$ & $\checkmark$ &              \\
Book Readability    &              & $\checkmark$ & $\checkmark$ &              \\
Company Size        &              & $\checkmark$ & $\checkmark$ &              \\
Montgomery Salaries &              & $\checkmark$ & $\checkmark$ &              \\
Ramen Ratings       &              & $\checkmark$ & $\checkmark$ &              \\
US Accidents        &              & $\checkmark$ & $\checkmark$ &              \\
Vancouver Salaries  &              & $\checkmark$ & $\checkmark$ &              \\
Zomato Restaurants  &              & $\checkmark$ & $\checkmark$ &              \\
California Prices   & $\checkmark$ &              &              & $\checkmark$ \\
Fake Job Postings   & $\checkmark$ &              &              & $\checkmark$ \\
Kickstarter         & $\checkmark$ &              &              & $\checkmark$ \\
Mercari Marketplace & $\checkmark$ &              &              & $\checkmark$ \\
Spotify Genres      &              & $\checkmark$ &              & $\checkmark$ \\
Beer Ratings        &              &              & $\checkmark$ & $\checkmark$ \\
\bottomrule
\end{tabular}
\end{table}

\subsection{Empirical Results for Curation Conditions}

\begin{figure*}[!htb]
\centering
\includegraphics[width=0.98\textwidth]{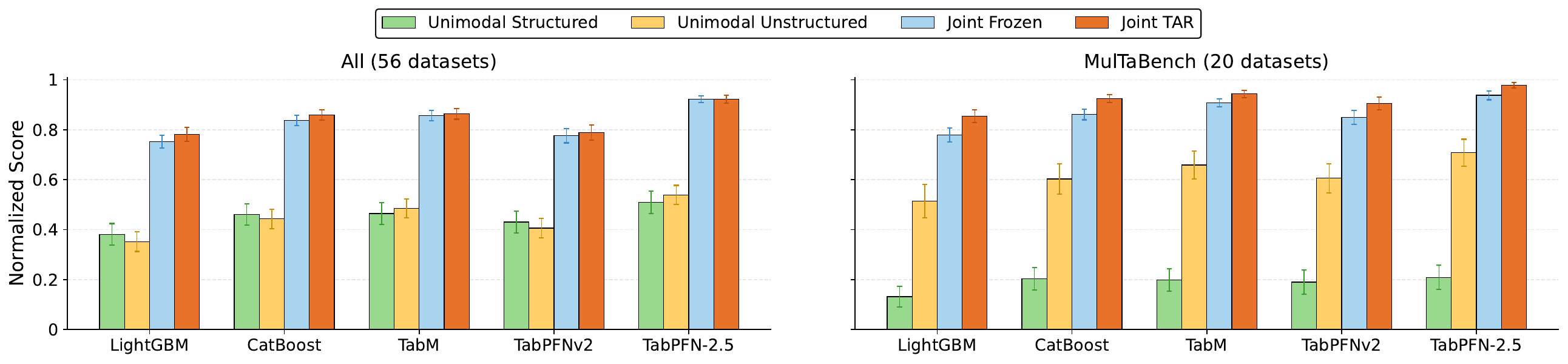}
\caption{Curations Conditions for the Text-Tabular Pool. Normalized scores for \textit{Structured}, \textit{Unstructured}, \textit{Joint Frozen}, and \textit{Joint TAR} across all 56 candidates (left) and the MulTaBench subset (right).}
\label{fig:text_pool_struct_unstruct}
\end{figure*}

Figure~\ref{fig:text_pool_struct_unstruct} shows normalized scores across all 4 conditions for the full pool and the MulTaBench subset. The \textit{Structured} and \textit{Unstructured} bars serve as unimodal baselines. The MulTaBench subset shows a consistent ordering across all 4 conditions, which is more pronounced than in the full corpus.

\subsection{Benchmark Acceptance Breakdown}

Table~\ref{tab:text_curation_rates} reports acceptance rates per source benchmark. Grinsztajn et al. and the AutoML Multimodal Benchmark yield the highest rates. CARTE has the lowest acceptance rate (33\%), reflecting its focus on knowledge-graph-style short strings and high-cardinality categorical columns. Out of 56 candidates, 23 pass all criteria; we retain 20 for MulTaBench.

\begin{table}[ht]
\centering
\caption{Text-tabular curation acceptance rates by source benchmark.}
\label{tab:text_curation_rates}
\small
\begin{tabular}{lrrr}
\toprule
\textbf{Benchmark} & \textbf{Candidates} & \textbf{Accepted} & \textbf{Rate} \\
\midrule
AutoML Multimodal  & 16 & 10 & 62\% \\
Grinsztajn et al.     & 11 &  7 & 64\% \\
CARTE        & 33 & 11 & 33\% \\
TextTabBench      & 13 &  7 & 54\% \\
\midrule
\textbf{Total (deduplicated)} & \textbf{56} & \textbf{23} & \textbf{41\%} \\
\bottomrule
\end{tabular}
\end{table}

\subsection{Per-Dataset Curation Results}\label{app:text_curation_grid}

Table~\ref{tab:text_curation_grid} reports, for each of the 56 candidate datasets, whether each of the five curation models satisfies both criteria jointly. A checkmark indicates both hold simultaneously; $\times$ indicates failure on at least one; $-$ indicates the model could not be evaluated, due to highly-multiclass problems where TabPFN's variants can't run on. Datasets are sorted approved-first, then by descending pass count.

\begin{longtable}{llcccccr}
\caption{Per-dataset curation grid. Models: LightGBM (LGBM), CatBoost (Cat), TabM, PFNv2 (TabPFNv2), PFN-2.5 (TabPFN-2.5). Each cell indicates whether the model satisfies criteria. \textit{Pass?} column shows how many models pass. Options are pass ($\checkmark$), fail ($\times$), and N/A ($-$).}
\label{tab:text_curation_grid}\\
\toprule
\textbf{Dataset} & \textbf{LGBM} & \textbf{Cat} & \textbf{TabM} & \textbf{PFNv2} & \textbf{PFN-2.5} & \textbf{Pass?} \\
\midrule
\endfirsthead
\multicolumn{8}{l}{\small\textit{(continued from previous page)}}\\[2pt]
\toprule
\textbf{Dataset} & \textbf{LGBM} & \textbf{Cat} & \textbf{TabM} & \textbf{PFNv2} & \textbf{PFN-2.5} & \textbf{Pass?} \\
\midrule
\endhead
\midrule
\multicolumn{8}{r}{\small\textit{(continued on next page)}}\\
\endfoot
\bottomrule
\endlastfoot
\multicolumn{8}{l}{\textit{Approved (23 datasets)}} \\[2pt]
Kickstarter Funding  & $\checkmark$ & $\checkmark$ & $\checkmark$ & $\checkmark$ & $\checkmark$ & 5 \\
Jigsaw Toxicity    & $\checkmark$ & $\checkmark$ & $\checkmark$ & $\checkmark$ & $\checkmark$ & 5 \\
Product Sentiment  & $\checkmark$ & $\checkmark$ & $\checkmark$ & $\checkmark$ & $\checkmark$ & 5 \\
Women's Clothing & $\checkmark$ & $\checkmark$ & $\checkmark$ & $\checkmark$ & $\checkmark$ & 5 \\
Michelin Guide  & $\checkmark$ & $\checkmark$ & $\checkmark$ & $\checkmark$ & $\checkmark$ & 5 \\
News Channel Category    & $\checkmark$ & $\checkmark$ & $\checkmark$ & $\checkmark$ & $\checkmark$ & 5 \\
Baby Products   & $\checkmark$ & $\checkmark$ & $\checkmark$ & $\checkmark$ & $\checkmark$ & 5 \\
Vancouver Salaries   & $\checkmark$ & $\checkmark$ & $\checkmark$ & $\checkmark$ & $\checkmark$ & 5 \\
SciMagojr Impact  & $\checkmark$ & $\checkmark$ & $\checkmark$ & $\checkmark$ & $\checkmark$ & 5 \\
Book Readability  & $\checkmark$ & $\checkmark$ & $\checkmark$ & $\checkmark$ & $\checkmark$ & 5 \\
Video Games Sales   & $\checkmark$ & $\checkmark$ & $\checkmark$ & $\checkmark$ & $\checkmark$ & 5 \\
Consumer Complaint   & $\checkmark$ & $\checkmark$ & $\checkmark$ & $\checkmark$ & $\times$     & 4 \\
Hearthstone Cards  & $\checkmark$ & $\checkmark$ & $\times$     & $\checkmark$ & $\checkmark$ & 4 \\
US Accidents   & $\checkmark$ & $\checkmark$ & $\times$     & $\checkmark$ & $\checkmark$ & 4 \\
Book Price  & $\checkmark$ & $\checkmark$ & $\checkmark$ & $\checkmark$ & $\times$     & 4 \\
Mercari Marketplace  & $\checkmark$ & $\checkmark$ & $\checkmark$ & $\checkmark$ & $\times$     & 4 \\
Zomato Restaurants   & $\checkmark$ & $\checkmark$ & $\checkmark$ & $\checkmark$ & $\times$     & 4 \\
Rotten Tomatoes   & $\checkmark$ & $\checkmark$ & $\times$     & $\checkmark$ & $\checkmark$ & 4 \\
Fake Job Posting   & $\checkmark$ & $\checkmark$ & $\times$     & $\checkmark$ & $\times$     & 3 \\
Wine Review  & $\checkmark$ & $\checkmark$ & $\checkmark$ & $-$          & $-$          & 3 \\
Data Scientist Salary  & $\times$     & $\checkmark$ & $\times$     & $\checkmark$ & $\checkmark$ & 3 \\
Spotify Genres   & $\checkmark$ & $\checkmark$ & $\checkmark$ & $-$          & $-$          & 3 \\
Montgomery Salaries  & $\times$     & $\times$     & $\checkmark$ & $\checkmark$ & $\checkmark$ & 3 \\
\midrule
\multicolumn{8}{l}{\textit{Rejected (33 datasets)}} \\[2pt]
OSHA Accident Injury   & $\checkmark$ & $\times$     & $\checkmark$ & $\times$     & $\times$     & 2 \\
Google Q\&A Type   & $\times$     & $\times$     & $\times$     & $\checkmark$ & $\checkmark$ & 2 \\
American Eagle Prices     & $\times$     & $\checkmark$ & $\checkmark$ & $\times$     & $\times$     & 2 \\
JC Penney Products   & $\checkmark$ & $\times$     & $\times$     & $\checkmark$ & $\times$     & 2 \\
Wikiliq Alcohol   & $\times$     & $\times$     & $\checkmark$ & $\checkmark$ & $\times$     & 2 \\
Chocolate Bar Ratings   & $\times$     & $\checkmark$ & $\checkmark$ & $\times$     & $\times$     & 2 \\
Wine Vivino Spain   & $\checkmark$ & $\times$     & $\checkmark$ & $\times$     & $\times$     & 2 \\
California House Prices  & $\checkmark$ & $\checkmark$ & $\times$     & $\times$     & $\times$     & 2 \\
SF Permit Applications   & $\times$     & $\times$     & $\checkmark$ & $\checkmark$ & $\times$     & 2 \\
FIFA22 Wages  & $\checkmark$ & $\times$     & $\times$     & $\checkmark$ & $\times$     & 2 \\
IMDB Genre  & $\times$     & $\times$     & $\times$     & $\checkmark$ & $\times$     & 1 \\
Melbourne Airbnb   & $\checkmark$ & $\times$     & $\times$     & $\times$     & $\times$     & 1 \\
Bike Price Bikewale   & $\times$     & $\times$     & $\times$     & $\checkmark$ & $\times$     & 1 \\
Car Price Cardekho   & $\checkmark$ & $\times$     & $\times$     & $\times$     & $\times$     & 1 \\
Polish Wine Prices  & $\times$     & $\checkmark$ & $\times$     & $\times$     & $\times$     & 1 \\
ML/DS Job Salaries  & $\times$     & $\times$     & $\checkmark$ & $\times$     & $\times$     & 1 \\
Books Goodreads    & $\times$     & $\checkmark$ & $\times$     & $\times$     & $\times$     & 1 \\
Korean Drama     & $\checkmark$ & $\times$     & $\times$     & $\times$     & $\times$     & 1 \\
US Museum Revenues   & $\checkmark$ & $\times$     & $\times$     & $\times$     & $\times$     & 1 \\
Used Cars Pakistan  & $\checkmark$ & $\times$     & $\times$     & $\times$     & $\times$     & 1 \\
Used Cars Saudi Arabia   & $\times$     & $\times$     & $\times$     & $\times$     & $\checkmark$ & 1 \\
Yelp Reviews    & $\times$     & $\times$     & $\times$     & $\times$     & $\times$     & 0 \\
Laptop Indian Prices   & $\times$     & $\times$     & $\times$     & $\times$     & $\times$     & 0 \\
Beer Ratings   & $\times$     & $\times$     & $\times$     & $\times$     & $\times$     & 0 \\
Coffee Review   & $\times$     & $\times$     & $\times$     & $\times$     & $\times$     & 0 \\
Ramen Ratings   & $\times$     & $\times$     & $\times$     & $\times$     & $\times$     & 0 \\
Airbnb Seattle   & $\times$     & $\times$     & $\times$     & $\times$     & $\times$     & 0 \\
Company Employee Size    & $\times$     & $\times$     & $\times$     & $\times$     & $\times$     & 0 \\
Anime Planet Rating   & $\times$     & $\times$     & $\times$     & $\times$     & $\times$     & 0 \\
FilmTV Movie Rating   & $\times$     & $\times$     & $\times$     & $\times$     & $\times$     & 0 \\
Movies Dataset Revenue    & $\times$     & $\times$     & $\times$     & $\times$     & $\times$     & 0 \\
NBA Draft VORP   & $\times$     & $\times$     & $\times$     & $\times$     & $\times$     & 0 \\
Mercedes Italy Cars   & $\times$     & $\times$     & $\times$     & $\times$     & $\times$     & 0 \\
\end{longtable}

\section{Image-Tabular Curation}
\label{app:image_curation}

\subsection{Existing Benchmarks}

The image-tabular benchmarking landscape is substantially more limited than its text-tabular counterpart. MuG~\cite{lu_mug_2023} reports 8 text-image-tabular datasets, but these correspond to only 4 underlying datasets, some of them using different target variables. \citet{tang_bag_2024} curate 22 datasets spanning varying modality combinations: 6 are text-tabular and overlap with text-tabular benchmarks; 5 are text-image datasets that lie outside the scope of this paper; and the remaining 11 qualify for our image-tabular definition (6 of them also have text). In addition, we include the datasets introduced by TIME \cite{luo_time_2025} and MultimodalTabPFN \cite{kim_multimodalpfn_2025}, some of them overlapping with aforementioned benchmarks. 

However, many of these datasets suffer from serious reproducibility problems. For example, the \href{https://www.kaggle.com/datasets/airbnb/seattle}{\textit{Seattle} dataset} contains links to images via external URLs that are no longer reachable; and the \textit{KARD} dataset points to a Kaggle dataset that has since been deleted. The remaining candidates are partially recoverable, but their preprocessing logic is often undocumented and difficult to replicate faithfully. 

After deduplication and removal of unavailable datasets, we are able to evaluate 16 unique datasets, from which only 5 pass the curation filter. For the ones which did non pass, it was sometimes hard to assess whether we have curated them properly. Therefore, we do not report curation statistics at the same level of detail as for text-tabular, and focus the remaining of the section on elaborating on the curation process.

\subsection{Curation Logic}

The curation of datasets found in the wild involved several decisions with the goal of making the image feature important and interesting enough to qualify as a relevant true image-tabular task.

\paragraph{Images.}
Each dataset contains exactly one image column; datasets with multiple image fields per row (e.g., product galleries) were reduced to a single image for simplicity. Rows with absent or corrupt image files are dropped without imputation, as there is no sensible substitute for a missing image, and placeholder images would inject noise into the encoding step.

\paragraph{Feature and Target engineering.}
In several cases the raw target required transformation before satisfying the curation criteria, and we provide a non-exhaustive list of examples. \textit{Log transformation}: Amazon Bestseller retail price is transformed as $\log(1 + \mathrm{price})$ to stabilize the regression target across several orders of magnitude. \textit{Quantile binning}: CS:GO Skin Price (10 equal-frequency bins), PetFinder listed age (8 bins), and HubMAP HPA donor age (10 bins) are discretized into multiclass targets. \textit{Feature removal}: structured columns that directly encode the target or fully dominate the image signal are dropped. This is particularly evident in examples like Zooscan Plankton, where features were extracted directly from the image, and removing them increased the image importance.

\paragraph{Kaggle upload.}
To ensure reproducibility, all the image-tabular datasets we release are preprocessed and uploaded to Kaggle under the MulTaBench organization. Each upload contains a flat \texttt{images/} directory with one file per row named consistently, a \texttt{data.csv} with features and target. The image column stores relative paths into \texttt{images/} directory. A unified loading API handles download and ingestion, ensuring all datasets are accessed identically regardless of original source format.

\section{Text-Image-Tabular Datasets}
\label{app:trimodal_datasets}

Within the 20 image-tabular datasets in MulTaBench, 9 include one or more text columns alongside the image and structured features. This section details our methodology for testing whether these datasets are genuinely trimodal, justifies our strict definition of trimodality, and explores the implications of adopting weaker criteria.

\subsection{A Strict Definition of Trimodality}

Consistent with our core motivation, the mere coexistence of text, images, and tabular data does not constitute a trimodal task unless each modality independently fulfills our strict multimodal criteria. We therefore apply the selection rule of Appendix~\ref{app:curation_formal} to each unstructured modality independently, and accept a dataset as trimodal only if all of the following hold for a majority of the curation learners:

\begin{enumerate}
\item \textbf{Joint Signal, per modality.} The joint frozen model must outperform both the individual unstructured modality and the bimodal baseline that omits it. Satisfying this for both text and images confirms that all three modalities provide complementary predictive signals.
\item \textbf{Task-awareness, per modality.} Applying TAR to either the image or the text individually improves upon the joint frozen condition, demonstrating that both frozen encoders discard task-relevant information.
\item \textbf{Joint Task-awareness.} Adapting both encoders simultaneously outperforms adapting either one alone, ensuring the dual adaptation is not redundant.
\end{enumerate}

These latter two conditions deliberately raise the bar compared to our bimodal requirements. Relaxing them would substantially increase the number of trimodal datasets in both the core benchmark and the extended collection. While we consider \textit{Joint Signal, per modality} to be strictly mandatory, as failing it implies a redundant modality, the \textit{Task-awareness} requirements could be more flexible. For instance, one could omit the \textit{Joint Task-awareness} constraint, or require \textit{Task-awareness} from only a single unstructured modality. Acknowledging that any strict binary cutoff is inherently rigid, we encourage future researchers to challenge our own definitions of trimodality.

\subsection{Per-Dataset Results}

Only 2 datasets from our pool of 9 candidates qualify, and they do so across all 5 learners for every condition. The remaining 7 qualify on no learner at all. The outcome is therefore insensitive to the consensus threshold $\rho$: any threshold perfectly separates the passing datasets from the failing ones, indicating that the hyperparameter choices are not what makes the trimodal subset small.

Interestingly, all candidates satisfy \textit{Joint Signal} for both modalities, and as expected, they all satisfy \textit{Task-awareness} for the image. The rejected datasets specifically fail \textit{Task-awareness} for the text: their text encoder, even when left frozen, already supplies the necessary predictive signal for the task.

The results for \textit{PetFinder} are presented in Table~\ref{tab:petfinder} of the main paper. \textit{Amazon Packages} is a regression task predicting the total weight of an Amazon bin from a warehouse photograph, a product description, and the expected item quantity. The results for this dataset are presented in Table~\ref{tab:amazon_packages}.

\begin{table}[ht]
\centering
\caption{Amazon Packages Analysis. S=Structured, I=Image, T=Text. Mean $R^2$ (\%) per model and condition. For all models, TAR over both modalities dominates.}
\label{tab:amazon_packages}
\setlength{\tabcolsep}{4pt}
\small
\begin{tabular}{l cc ccc ccc}
\toprule
 & \multicolumn{2}{c}{\textit{Single modality}} & \multicolumn{3}{c}{\textit{Frozen combinations}} & \multicolumn{3}{c}{\textit{Target-Aware Representations (TAR)}} \\
\cmidrule(lr){2-3}\cmidrule(lr){4-6}\cmidrule(lr){7-9}
\textbf{Model} & I & T & S+I & S+T & S+I+T & S+I\textsubscript{TAR}+T & S+I+T\textsubscript{TAR} & \textbf{S+I\textsubscript{TAR}+T\textsubscript{TAR}} \\
\midrule
LightGBM   & 43.2 & 17.5 & 45.4 & 20.8 & 49.9 & 56.9 & 52.2 & \textbf{59.4} \\
CatBoost   & 44.2 & 19.0 & 46.7 & 22.5 & 52.5 & 58.1 & 53.8 & \textbf{59.8} \\
TabM       & 46.6 & 20.5 & 48.6 & 24.2 & 55.7 & 60.3 & 56.1 & \textbf{61.1} \\
TabPFNv2   & 46.6 & 20.2 & 49.3 & 23.9 & 54.9 & 59.8 & 56.2 & \textbf{61.3} \\
TabPFN-2.5 & 47.5 & 21.1 & 50.2 & 24.9 & 56.2 & 60.7 & 57.2 & \textbf{61.5} \\
\bottomrule
\end{tabular}
\end{table}

\subsection{Additional Trimodal Candidates}

Among the 20 supplementary image-tabular datasets in Appendix~\ref{app:extra_datasets}, 13 also carry text columns. Across the entire 80-dataset release, 22 datasets thus feature all three modalities. Under our strict criteria, these 13 datasets are ineligible for trimodal status because most already fail \textit{Task-awareness} for the image encoder, rendering a complete text-modality curation redundant. We therefore leave these candidates uncurated for text, noting that they offer a natural starting point for future work developing dedicated trimodal tabular benchmarks.

\section{Additional Released Datasets}
\label{app:extra_datasets}

Alongside the 40 MulTaBench datasets we release 40 further datasets, 20 image-tabular and 20 text-tabular, bringing the released collection to 80 datasets of which 40 are image-tabular. All of them satisfy the \textit{Joint Signal} criterion, but do not necessarily meet the \textit{Task-awareness} criterion. Most of them fail it, and others failed it under a different curation regime; a few pass it weakly and were excluded to keep the benchmark at exactly 40 datasets, balanced between the image and text modalities and between classification and regression.

These additional datasets are therefore intended for work on multimodal fusion, where the question is how to combine modalities rather than how to adapt the encoder. Table~\ref{tab:extra_datasets} reports the per-dataset scores behind their admission and Table~\ref{tab:extra_datasets_properties} their properties, with the two subsections below describing each dataset individually, following the format used for MulTaBench in Appendix~\ref{app:multabench}.

\subsection{Image-Tabular Dataset Descriptions}

\begin{table*}[!tp]
\centering
\caption{Datasets released alongside MulTaBench, admitted on \textit{Joint Signal}. Structured, Unstructured and Joint are the mean score over the curation learners under each condition, AUC for classification and $R^2$ for regression. $\Delta_{\text{Joint}}$ is the median of Joint minus the better unimodal condition, and Pass counts the learners for which it reaches $\delta$, out of those evaluated.}
\label{tab:extra_datasets}
\setlength{\tabcolsep}{5pt}\small
\begin{tabular}{l c r r r r c}
\toprule
\textbf{Dataset} & Task & Structured & Unstructured & Joint & $\Delta_{\text{Joint}}$ & Pass \\
\midrule
\multicolumn{7}{c}{\textit{Image-Tabular}} \\
\midrule
Watch Tier           & REG & 0.341 & 0.201 & 0.479 & 0.150 & 5/5 \\
eMAG Products        & REG & 0.749 & 0.769 & 0.841 & 0.063 & 5/5 \\
Zepto Groceries      & REG & 0.670 & 0.467 & 0.727 & 0.054 & 5/5 \\
Tokopedia Weight     & REG & 0.635 & 0.422 & 0.687 & 0.040 & 5/5 \\
PAD-UFES Lesions     & CLS & 0.921 & 0.896 & 0.950 & 0.030 & 5/5 \\
Minecraft Mobs       & CLS & 0.940 & 0.957 & 0.988 & 0.028 & 5/5 \\
Pinterest Repins     & CLS & 0.622 & 0.638 & 0.659 & 0.025 & 5/5 \\
Airbnb NYC           & REG & 0.701 & 0.309 & 0.721 & 0.020 & 5/5 \\
OASIS Alzheimer's    & CLS & 0.910 & 0.893 & 0.931 & 0.019 & 5/5 \\
SoCal Houses         & REG & 0.735 & 0.339 & 0.757 & 0.017 & 5/5 \\
Goias Apartments     & REG & 0.931 & 0.660 & 0.947 & 0.015 & 5/5 \\
Kamernet Room Size   & REG & 0.446 & 0.129 & 0.459 & 0.015 & 5/5 \\
Sao Paulo Apartments & REG & 0.833 & 0.420 & 0.847 & 0.015 & 5/5 \\
Lahaina Art Auction  & REG & 0.558 & 0.093 & 0.576 & 0.013 & 5/5 \\
Reddit Memes         & CLS & 0.929 & 0.766 & 0.939 & 0.013 & 4/5 \\
Flipkart Discount    & REG & 0.710 & 0.450 & 0.722 & 0.012 & 5/5 \\
DVM Car Prices       & REG & 0.592 & 0.357 & 0.594 & 0.009 & 4/5 \\
Hearthstone Class    & CLS & 0.905 & 0.812 & 0.914 & 0.009 & 3/3 \\
Pokemon Height       & CLS & 0.849 & 0.709 & 0.853 & 0.007 & 4/5 \\
HAM10000 Lesions     & CLS & 0.798 & 0.949 & 0.956 & 0.005 & 5/5 \\
\cmidrule(lr){3-6}
\textit{Mean}        & & 0.739 & 0.562 & 0.777 & 0.016 & \\
\midrule
\multicolumn{7}{c}{\textit{Text-Tabular}} \\
\midrule
Chocolate Bars       & REG & 0.166 & 0.130 & 0.257 & 0.104 & 5/5 \\
Wine Poland          & REG & 0.458 & 0.159 & 0.539 & 0.087 & 4/5 \\
Anime Planet         & REG & 0.661 & 0.616 & 0.751 & 0.086 & 5/5 \\
WikiLiq Spirits      & REG & -0.001 & 0.155 & 0.226 & 0.060 & 5/5 \\
Used Cars Pakistan   & REG & 0.863 & 0.820 & 0.911 & 0.050 & 5/5 \\
Box Office           & CLS & 0.819 & 0.763 & 0.861 & 0.042 & 5/5 \\
Goodreads Books      & CLS & 0.706 & 0.724 & 0.756 & 0.031 & 5/5 \\
Used Cars Saudi      & REG & 0.867 & 0.174 & 0.895 & 0.029 & 5/5 \\
Consumer Complaints  & CLS & 0.848 & 0.887 & 0.906 & 0.026 & 5/5 \\
Ramen Ratings        & REG & 0.087 & 0.159 & 0.181 & 0.025 & 5/5 \\
FIFA22 Wages         & REG & 0.761 & 0.042 & 0.775 & 0.017 & 5/5 \\
IMDB Genre           & CLS & 0.832 & 0.753 & 0.844 & 0.014 & 5/5 \\
Wine Vivino Spain    & REG & 0.816 & 0.773 & 0.826 & 0.011 & 5/5 \\
OSHA Injury          & CLS & 0.580 & 0.573 & 0.592 & 0.010 & 4/5 \\
Korean Drama         & CLS & 0.640 & 0.694 & 0.702 & 0.008 & 5/5 \\
Airbnb Seattle       & REG & 0.649 & 0.454 & 0.660 & 0.008 & 4/5 \\
News Channel         & CLS & 0.802 & 0.609 & 0.809 & 0.007 & 5/5 \\
American Eagle       & CLS & 0.994 & 0.989 & 0.996 & 0.003 & 4/5 \\
California Prices    & CLS & 0.990 & 0.956 & 0.993 & 0.002 & 5/5 \\
Airbnb Melbourne     & CLS & 0.838 & 0.779 & 0.837 & 0.002 & 3/5 \\
\cmidrule(lr){3-6}
\textit{Mean}        & & 0.669 & 0.560 & 0.716 & 0.021 & \\
\bottomrule
\end{tabular}
\end{table*}
\begin{table*}[!tp]
\centering
\caption{All 40 Additional Released Datasets Properties. \textit{Task}: Classification (CLS) or Regression (REG). \textit{Classes}: number of target classes (for CLS). \textit{N}: total examples. \textit{Struct.}: numerical + categorical features. \textit{Text}: free-text features. \textit{Img.}: image features.}
\label{tab:extra_datasets_properties}
\footnotesize
\begin{tabular}{llrrrrr}
\toprule
Dataset & Task & Classes & $N$ & Struct. & Text & Img. \\
\midrule
\multicolumn{7}{l}{\textit{Image-Tabular (20 datasets)}} \\
\midrule
HAM10000 Lesions     & CLS &   7 &   7,470 &   3 &  0 & 1 \\
Hearthstone Class    & CLS &  11 &   4,281 &  11 &  4 & 1 \\
Minecraft Mobs       & CLS &   7 &   6,089 &   7 &  0 & 1 \\
OASIS Alzheimer's    & CLS &   2 &     436 &   7 &  0 & 1 \\
PAD-UFES Lesions     & CLS &   6 &   1,641 &  21 &  0 & 1 \\
Pinterest Repins     & CLS &   2 &   1,504 &  15 &  0 & 1 \\
Pokemon Height       & CLS &  10 &   1,077 &  27 &  4 & 1 \\
Reddit Memes         & CLS &   7 &   4,821 &   6 &  0 & 1 \\
Airbnb NYC           & REG &  -- &  16,780 &  46 &  3 & 1 \\
DVM Car Prices       & REG &  -- & 100,000 &  14 &  1 & 1 \\
eMAG Products        & REG &  -- &  38,294 &   2 &  0 & 1 \\
Flipkart Discount    & REG &  -- &  60,900 &   1 &  2 & 1 \\
Goias Apartments     & REG &  -- &   9,830 &  73 &  1 & 1 \\
Kamernet Room Size   & REG &  -- &   1,638 &  20 &  4 & 1 \\
Lahaina Art Auction  & REG &  -- &     968 &   6 &  2 & 1 \\
Sao Paulo Apartments & REG &  -- &  23,432 &  28 &  1 & 1 \\
SoCal Houses         & REG &  -- &  15,474 &   4 &  2 & 1 \\
Tokopedia Weight     & REG &  -- &   1,197 &  13 &  2 & 1 \\
Watch Tier           & REG &  -- &   1,260 &   1 &  1 & 1 \\
Zepto Groceries      & REG &  -- &  13,345 &   5 &  2 & 1 \\
\midrule
\multicolumn{7}{l}{\textit{Text-Tabular (20 datasets)}} \\
\midrule
Airbnb Melbourne     & CLS &  10 &  18,316 &  68 & 20 & 0 \\
American Eagle       & CLS &   5 &  22,662 &   8 &  5 & 0 \\
Box Office           & CLS &   5 &   7,408 &  10 &  9 & 0 \\
California Prices    & CLS &   2 &  37,951 &  21 & 18 & 0 \\
Consumer Complaints  & CLS &   8 & 100,000 &  11 &  5 & 0 \\
Goodreads Books      & CLS &   3 &   3,967 &   6 &  8 & 0 \\
IMDB Genre           & CLS &   2 &     800 &   7 &  4 & 0 \\
Korean Drama         & CLS &  10 &   1,647 &   3 &  6 & 0 \\
News Channel         & CLS &   6 &  20,284 &  16 &  1 & 0 \\
OSHA Injury          & CLS &   2 &   4,847 &  13 &  3 & 0 \\
Airbnb Seattle       & REG &  -- &   3,818 &  51 & 14 & 0 \\
Anime Planet         & REG &  -- &  14,391 &   7 &  7 & 0 \\
Chocolate Bars       & REG &  -- &   1,795 &   5 &  3 & 0 \\
FIFA22 Wages         & REG &  -- &  19,178 &  27 &  1 & 0 \\
Ramen Ratings        & REG &  -- &   4,105 &   2 &  2 & 0 \\
Used Cars Pakistan   & REG &  -- &  72,655 &   6 &  3 & 0 \\
Used Cars Saudi      & REG &  -- &   8,035 &  11 &  1 & 0 \\
WikiLiq Spirits      & REG &  -- &  12,569 &   6 &  6 & 0 \\
Wine Poland          & REG &  -- &   2,247 &  13 &  5 & 0 \\
Wine Vivino Spain    & REG &  -- &   8,650 &   5 &  2 & 0 \\
\bottomrule
\end{tabular}
\end{table*}

\paragraph{\href{https://www.kaggle.com/datasets/kmader/skin-cancer-mnist-ham10000}{HAM10000 Lesions.}} Dermatoscopic images of common pigmented skin lesions gathered at two sites over twenty years.
The 7-class target is the diagnosis, covering melanoma, melanocytic nevi, basal cell carcinoma, actinic keratoses, benign keratosis-like lesions, dermatofibroma, and vascular lesions.
Structured features include the patient age, sex, and the anatomical localization of the lesion; the diagnostic confirmation method is dropped because it leaks the label.

\paragraph{\href{https://figshare.com/ndownloader/files/38561075}{Hearthstone Class.}} Collectible cards from the digital card game Hearthstone, paired with their illustration art.
The 11-class target is the hero class the card belongs to, such as Mage, Druid, or Warrior.
Structured features include the card statistics: mana cost, attack, health, rarity, card type, race, and set, while the free-text features cover the card name, rules text, mechanics list, and illustrator credit.

\paragraph{\href{https://www.kaggle.com/datasets/sqdartemy/minecraft-screenshots-dataset-with-features}{Minecraft Mobs.}} Screenshots of Minecraft gameplay annotated with the game state at the moment of capture.
The 7-class target is the mob the player is facing.
Structured features record the player's activity, remaining hearts, light level, and item in hand, together with the decision fields derived from them; the column restating the mob label is dropped.

\paragraph{\href{https://www.kaggle.com/datasets/shreyanmohanty/oasis-alzheimers-detection-multi-class-dataset}{OASIS Alzheimer's.}} Cross-sectional brain MRI scans from the Open Access Series of Imaging Studies.
The binary target is whether the subject is demented.
Structured features include the sex, age, years of education, socioeconomic status, estimated total intracranial volume, atlas scaling factor, and scan delay; clinical gradings such as CDR and MMSE are dropped because they restate the diagnosis.

\paragraph{\href{https://www.kaggle.com/datasets/mahdavi1202/skin-cancer}{PAD-UFES Lesions.}} Smartphone photographs of skin lesions collected in Brazil by the Dermatological and Surgical Assistance Program.
The 6-class target is the diagnosis, spanning basal and squamous cell carcinoma, melanoma, actinic and seborrheic keratosis, and nevus.
Structured features cover patient demographics and habits, family cancer history, environmental exposures, Fitzpatrick skin type, lesion region and size, and symptoms such as itching, bleeding, and growth; the biopsy indicator is dropped as it leaks the label.

\paragraph{\href{https://www.kaggle.com/datasets/andreacombette/pinterest-analysis-using-nlp-and-image-analysis}{Pinterest Repins.}} Pinterest pins paired with their images and engagement counts.
The binary target is whether the pin is widely repinned, defined by a median split of the repin count.
The pin title and description are compressed into E5 components, which form the structured features.

\paragraph{\href{https://www.kaggle.com/datasets/divyanshusingh369/complete-pokemon-library-32k-images-and-csv}{Pokemon Height.}} Sprites from the Pokemon franchise, paired with the game statistics of each entry.
The 10-class target is the height decile, obtained after capping height at 10 meters.
Structured features include the base, minimum, and maximum battle statistics together with weight, catch rate, base friendship and experience, growth rate, egg groups, and gender ratio, while the text features cover the name, species, type, and ability list.

\paragraph{\href{https://www.kaggle.com/datasets/musadiqpashak/reddit-memes-dataset}{Reddit Memes.}} Meme images scraped from Reddit together with the posting account and its engagement statistics.
The 7-class target is the subreddit category the meme was posted to.
Structured features include the poster's total and comment karma, the upvote ratio, the number of comments, the time of day, and the account cake day; the title, extracted caption, author, and upvote count are dropped.

\paragraph{\href{https://www.kaggle.com/datasets/dominoweir/inside-airbnb-nyc}{Airbnb NYC.}} Airbnb listings in New York City paired with the primary listing photograph, downscaled to a maximum side of 512 pixels.
The regression target is the nightly price.
Structured features cover host response and acceptance, superhost status, property and room type, capacity, stay limits, availability, and review counts and scores, with the neighbourhood and host location as text features; host prose, scrape identifiers, and coordinates are dropped.

\paragraph{\href{https://www.kaggle.com/datasets/osamasaifs/dvm-car-with-images}{DVM Car Prices.}} Advertisement records from the DVM-CAR collection of the UK used car market, each paired with a photograph of the vehicle.
The regression target is the advertised price.
Structured features include the maker, advertisement date, colour, registration year, body type, mileage, engine size, gearbox, fuel type, seat and door counts, the predicted viewpoint of the photograph, and a quality check flag, with the general model name as a text feature.

\paragraph{\href{https://www.kaggle.com/datasets/furduisorinoctavian/romanian-products-from-emag-and-flanco}{eMAG Products.}} Product listings scraped from the Romanian retailers eMAG and Flanco, each with a product photograph.
The regression target is the price.
Only the customer rating and the product category are kept as structured features; the Romanian product title, identifiers, and URLs are dropped because the title names the object shown in the image.

\paragraph{\href{https://www.kaggle.com/datasets/kuchhbhi/stylish-product-image-dataset}{Flipkart Discount.}} Fashion product listings from Flipkart paired with their catalogue photographs.
The regression target is the discount ratio, the sold price divided by the listed price.
The listed price is the only structured feature, with the brand and the product title as text features; the sold price is dropped as it is the numerator of the target.

\paragraph{\href{https://www.kaggle.com/datasets/carloseduardogo/apartment-prices-in-goinia-gois-brazil}{Goias Apartments.}} Apartment listings from Goiania in the Brazilian state of Goias, each with a photograph of the unit.
The regression target is the asking price.
Structured features include the floor area, room, bathroom and parking counts, floor level, listing stage and business type, and binary amenity flags such as pool, elevator, and gym, with the neighbourhood as a text feature; the Portuguese description, address, condominium fee, and property tax are dropped.

\paragraph{\href{https://www.kaggle.com/datasets/amancapy/all-kamernet-rent-listings-w-photos-netherlands}{Kamernet Room Size.}} Room rental listings from the Dutch platform Kamernet, each with a photograph of the room.
The regression target is the floor area of the room in square meters.
Structured features include the place and furnishing type, rent and additional costs, tenant and roommate composition, shared facility flags, the energy label, and the availability start date, with the city, requested age range, and listed extras as text features; the listing description is dropped because it states the floor area.

\paragraph{\href{https://www.kaggle.com/datasets/quillen/artists-for-lahaina-2023}{Lahaina Art Auction.}} Artworks offered in the 2023 Artists for Lahaina charity auction, each with a photograph of the piece.
The regression target is the winning bid.
Structured features include the medium, the canvas dimensions, the stated value, the buy-now price, and the gallery, with the artist name and the work title as text features.

\paragraph{\href{https://www.kaggle.com/datasets/rogerio/image-based-property-price-prediction-in-so-paulo}{Sao Paulo Apartments.}} Property listings from Sao Paulo paired with a photograph of the unit.
The regression target is the asking price.
Structured features include the listing date and featured flag, property category and type, clipped floor area, bedroom, bathroom and parking counts, zone, and a set of amenity flags, with the neighbourhood as a text feature; the Portuguese headers and values were translated to English, and the previous price, condominium fee, and property tax were dropped.

\paragraph{\href{https://www.kaggle.com/datasets/ted8080/house-prices-and-images-socal}{SoCal Houses.}} House listings from Southern California paired with a photograph of the property.
The regression target is the price.
Structured features include the city identifier, the bedroom and bathroom counts, and the floor area in square feet, with the street address and the city name as text features.

\paragraph{\href{https://www.kaggle.com/datasets/nsmlehq/tokopedia-products-2025}{Tokopedia Weight.}} Product listings from the Indonesian marketplace Tokopedia paired with their catalogue photographs.
The regression target is the shipping weight of the product.
Structured features include the order quantity limits, condition and status, price and discount, shop city, transaction counts, units sold, review count, rating, and number of images, with the category and shop names as text features; identifiers, URLs, and the Indonesian title and description are dropped.

\paragraph{\href{https://www.kaggle.com/datasets/mathewkouch/a-dataset-of-watches}{Watch Tier.}} Product photographs of wristwatches spanning fifty brands.
The regression target is the price of the watch relative to the median price of its brand, so that the label measures within-brand positioning rather than brand identity.
The brand is the only structured feature, with the model name as a text feature.

\paragraph{\href{https://www.kaggle.com/datasets/simpleaditya/zepto-products-dataset}{Zepto Groceries.}} Grocery listings from the Indian quick-commerce service Zepto paired with their product images.
The regression target is the price.
Structured features include the rating, review count, stock status, category, and storefront interface, with the pack quantity and the sub-category as text features; the undiscounted price and the product name are dropped.

\subsection{Text-Tabular Dataset Descriptions}

\paragraph{\href{https://www.openml.org/search?type=data&id=46665}{Airbnb Melbourne.}} Airbnb listings in Melbourne.
The 10-class target is the nightly price binned into deciles.
Free-text features include the host's descriptions of the listing, the space, the neighbourhood, transit, access, interaction, and house rules, together with the amenity list, while the structured features cover host attributes, property type and capacity, availability, review scores, and host verification flags.

\paragraph{\href{https://www.openml.org/search?type=data&id=46656}{American Eagle.}} Apparel products from the American Eagle catalogue.
The 5-class target is the price binned into quintiles.
Free-text features include the product name, description, style attributes, available sizes, and the product page URL, while the structured features cover the recommended retail price, brand, category, retailer, rating, review count, and colour.

\paragraph{\href{https://www.kaggle.com/datasets/rounakbanik/the-movies-dataset}{Box Office.}} Films from The Movies Dataset, restricted to titles that report a non-zero box office revenue.
The 5-class target is the revenue binned into quintiles.
Free-text features include the title, overview, tagline, genres, collection, production companies and countries, and spoken languages, while the structured features cover the budget, runtime, popularity, vote statistics, original language, release date, and the adult and video flags.

\paragraph{\href{https://www.openml.org/search?type=data&id=46669}{California Prices.}} Residential property listings from California.
The binary target is whether the sold price is above the median.
Free-text features include the address, the listing summary, the heating, cooling, parking, flooring, laundry, and appliance descriptions, and the school names, while the structured features cover the year built, room counts, interior area, school scores and distances, tax assessment, and previous prices.

\paragraph{\href{https://www.kaggle.com/datasets/selener/consumer-complaint-database}{Consumer Complaints.}} Complaints filed with the US Consumer Financial Protection Bureau.
The 8-class target is the company's response to the consumer, such as closed with explanation or closed with monetary relief.
Free-text features include the consumer complaint narrative, the issue, and the company name, while the structured features cover the product, state, submission channel, receipt and dispatch dates, and the timely response and dispute flags.

\paragraph{\href{http://pages.cs.wisc.edu/~anhai/data/784_data/books2/csv_files/goodreads.csv}{Goodreads Books.}} Books scraped from Goodreads.
The 3-class target is the average rating binned into tertiles.
Free-text features include the title, description, publisher, author names, and ISBN, while the structured features cover the page count, number of ratings, format, language, and publication date.

\paragraph{\href{https://www.openml.org/search?type=data&id=46667}{IMDB Genre.}} Films from the IMDB top-rated list.
The binary target is whether the film is a drama.
Free-text features include the title, plot description, director, and cast list, while the structured features cover the rank, year, runtime, rating, vote count, revenue, and Metascore.

\paragraph{\href{https://www.kaggle.com/datasets/noorrizki/top-korean-drama-list-1500}{Korean Drama.}} Top-rated Korean television dramas.
The 10-class target is the viewer score binned into deciles.
Free-text features include the title, synopsis, genre, tags, main cast, and broadcast network, while the structured features cover the year, content rating, and number of episodes.

\paragraph{\href{https://www.openml.org/search?type=data&id=46652}{News Channel.}} Online news articles described by the surface statistics of their body text.
The 6-class target is the publishing channel: World, Tech, Entertainment, Business, Social Media, or Lifestyle.
The article title is the only text feature, while the structured features include token counts and ratios, link, image and video counts, average token length, subjectivity, sentiment polarity, and positive and negative word rates.

\paragraph{\href{https://www.kaggle.com/datasets/ruqaiyaship/osha-accident-and-injury-data-1517}{OSHA Injury.}} Accident and injury reports filed with the US Occupational Safety and Health Administration between 2015 and 2017.
The binary target is whether the injured worker was regularly assigned to the task being performed.
Free-text features include the accident abstract, event description, and event keywords, while the structured features describe the event date, the work setting, the project cost and type, the degree and nature of the injury, and the contributing factors.

\paragraph{\href{https://www.kaggle.com/datasets/airbnb/seattle}{Airbnb Seattle.}} Airbnb listings in Seattle.
The regression target is the nightly price.
Free-text features include the host's descriptions of the listing, space, neighbourhood, transit, and house notes, the host's own profile text, and the amenity and verification lists, while the structured features cover host attributes, location, property type and capacity, availability, review scores, and booking policies.

\paragraph{\href{https://www.kaggle.com/datasets/hernan4444/animeplanet-recommendation-database-2020}{Anime Planet.}} Anime titles scraped from Anime-Planet.
The regression target is the average rating score.
Free-text features include the name and alternative name, the synopsis, tags, content warnings, studios, and season, while the structured features cover the vote count, type, episode count, duration, and the start and end years.

\paragraph{\href{https://www.kaggle.com/datasets/rtatman/chocolate-bar-ratings}{Chocolate Bars.}} Expert reviews of plain dark chocolate bars.
The regression target is the reviewer's rating.
Free-text features include the manufacturer name, the specific bean origin or bar name, and the broad bean origin, while the structured features cover the review reference number, review date, cocoa percentage, company location, and bean type.

\paragraph{\href{https://www.openml.org/search?type=data&id=45012}{FIFA22 Wages.}} Player records from the FIFA 22 video game.
The regression target is the weekly wage in euros.
The nationality is the only text feature, while the structured features include the age, height and weight, the overall and potential ratings, and skill ratings covering attacking, dribbling, passing, movement, defending, and goalkeeping.

\paragraph{\href{https://www.kaggle.com/datasets/ankanhore545/top-ramen-ratings-2022}{Ramen Ratings.}} Instant ramen products reviewed by The Ramen Rater.
The regression target is the star rating.
The brand and the variety name are the text features, with the packaging style and the country of origin as the only structured features.

\paragraph{\href{https://www.kaggle.com/datasets/mustafaimam/used-car-prices-in-pakistan-2021}{Used Cars Pakistan.}} Used car listings from the Pakistani market in 2021.
The regression target is the asking price.
Free-text features include the model, version, and registration city, while the structured features cover the make, manufacture year, engine displacement, assembly, mileage, and transmission.

\paragraph{\href{https://www.kaggle.com/datasets/turkibintalib/saudi-arabia-used-cars-dataset}{Used Cars Saudi.}} Used car listings from Saudi Arabia.
The regression target is the asking price.
The car type is the only text feature, while the structured features include the make, year, origin, colour, options package, engine size, fuel type, gear type, mileage, region, and whether the price is negotiable.

\paragraph{\href{https://www.kaggle.com/datasets/limtis/wikiliq-dataset}{WikiLiq Spirits.}} Spirit products listed on the WikiLiq price aggregator.
The regression target is the price.
Free-text features include the product name, brand, country, category list, tasting notes, and description, while the structured features cover the alcohol by volume, base ingredient, years aged, rating, rating count, and bottle volume.

\paragraph{\href{https://www.kaggle.com/datasets/skamlo/wine-price-on-polish-market}{Wine Poland.}} Wines offered on the Polish retail market.
The regression target is the price in Polish zloty.
Free-text features include the wine name, region, appellation, vineyard, and grape varieties, while the structured features cover the country, vintage, bottle volume, alcohol percentage, serving temperature, colour, kind, taste, style, medals, vegan and natural flags, and the critic score.

\paragraph{\href{https://www.kaggle.com/datasets/joshuakalobbowles/vivino-wine-data}{Wine Vivino Spain.}} Spanish wines listed on Vivino.
The regression target is the user rating.
The wine name and the winery are the text features, while the structured features include the vintage year, number of reviews, price, and region.

\section{Extended Results}
\label{app:extended_results}

\subsection{Main Results Breakdown}\label{app:results_breakdown}

\paragraph{New Models.} We extend our models suite by adding new models. 

\begin{itemize}
\item For XGBoost, we follow previous work \cite{gorishniy_revisiting_2021, arazi_tabstar_2025} and use the default implementation from the \textit{xgboost} package,\footnote{\url{https://pypi.org/project/xgboost/}} with $booster = gbtree, early\_stopping\_rounds = 50, n\_estimators = 2000$. 

\item For RandomForest, we use the default scikit-learn implementation with default configuration with $n\_estimators = 100$. 

\item For \textit{RealMLP} we use its official implementation in the \textit{pytabkit} package, disable label smoothing and optimize for cross\_entropy for binary classification and
$1-auc\_ovr$ for multiclass classification, keeping the other default hyperparameters. 

\item For \textit{TabICLv2,}\footnote{\url{https://pypi.org/project/tabicl/}} \textit{TabSTAR,}\footnote{\url{https://pypi.org/project/tabstar/}} \textit{ConTextTab,}\footnote{\url{https://github.com/SAP-samples/sap-rpt-1-oss}} and \textit{TabDPT,}\footnote{\url{https://pypi.org/project/tabdpt/}} we use their default implementations.

\item For \textit{AutoGluon-Multimodal} we use \textit{MultiModalPredictor}\footnote{\url{https://auto.gluon.ai/stable/api/autogluon.multimodal.MultiModalPredictor.html}} with $pretrained=True$ and optimizing for $roc\_auc$ (binary classification), $roc\_auc\_ovr$ (multiclass classification), and $r^2$ (regression).
\end{itemize}

\begin{figure*}[ht]
    \centering
    \includegraphics[width=0.98\textwidth]{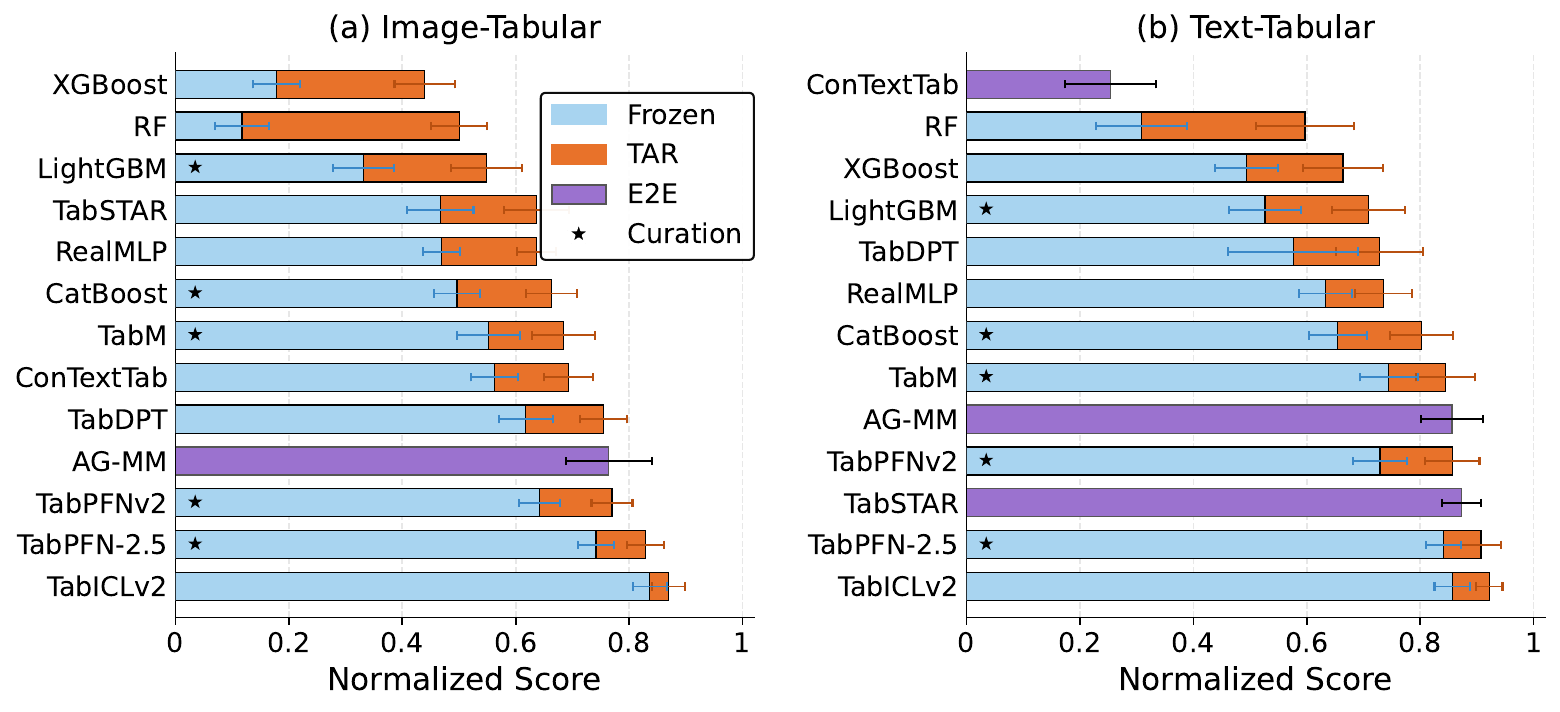}
    \caption{Tabular Learners Performances Analysis for Classification Tasks. Normalized scores over MulTaBench, with $\pm$ 95\% CI.}
    \label{fig:leaderboard_cls}
\end{figure*}

\begin{figure*}[ht]
    \centering
    \includegraphics[width=0.98\textwidth]{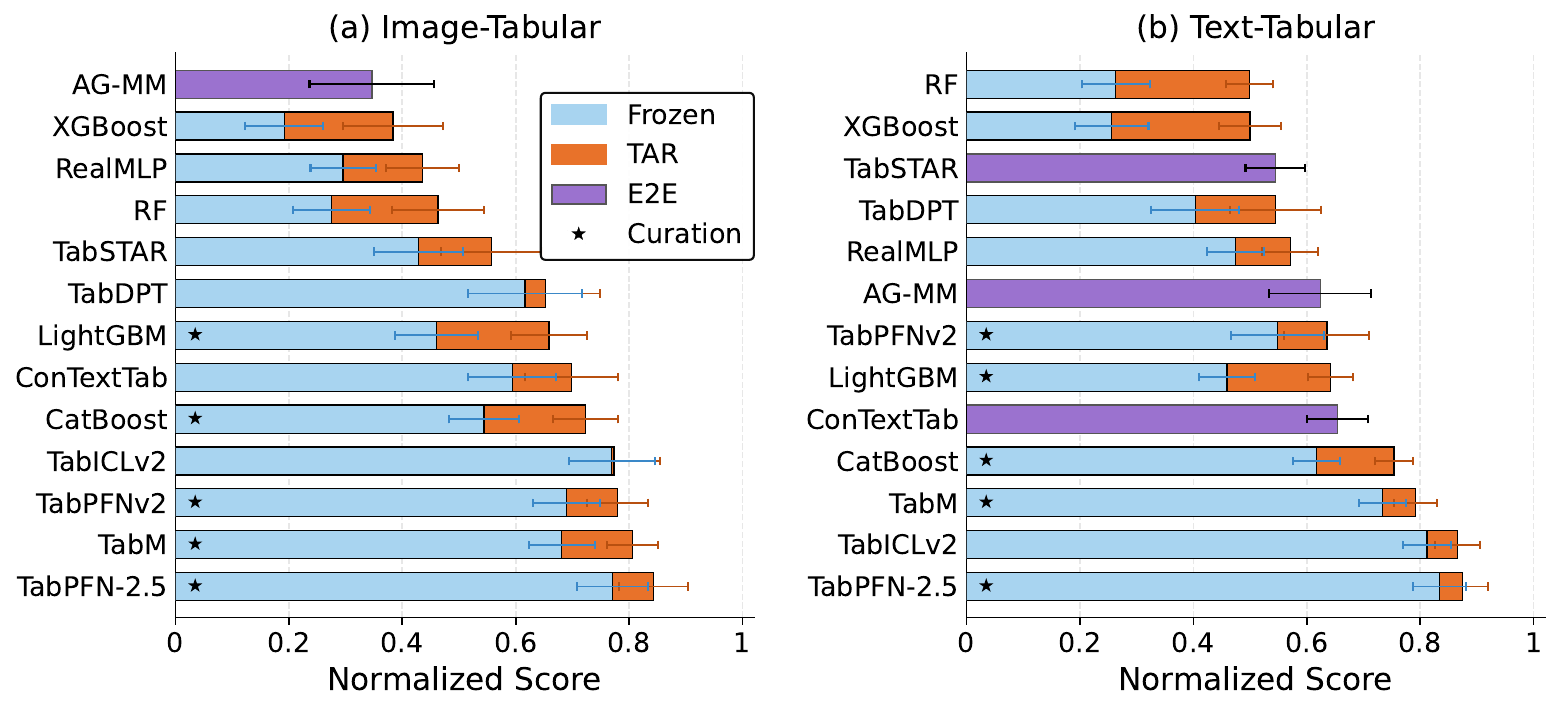}
    \caption{Tabular Learners Performances Analysis for Regression Tasks. Normalized scores over MulTaBench, with $\pm$ 95\% CI.}
    \label{fig:leaderboard_reg}
\end{figure*}

\paragraph{Task Type Breakdown} 
Figures~\ref{fig:leaderboard_cls} and~\ref{fig:leaderboard_reg} replicate Figure~\ref{fig:leaderboard}, but breaking down to classification and regression datasets respectively.
TAR consistently outperforms Frozen in both task types and both modalities, indicating that the benefit of target-aware representations is not specific to any of them.

\paragraph{Win Rate by Model} Table~\ref{tab:win_rate} reports the fraction of (dataset, fold) pairs where TAR outperforms Frozen for each model, with 95\% CIs.
End-to-end systems that do not expose a separate TAR condition for a given modality (TabSTAR, ConTextTab) are excluded from the corresponding column.
TAR beats Frozen in the large majority of runs across all models and both modalities.

\begin{table}[ht]
\centering
\caption{Per-model TAR win rate on MulTaBench. End-to-end models excluded from columns where they lack a separate TAR condition.}
\label{tab:win_rate}
\small
\begin{tabular}{lccc}
\toprule
Model & Image (\%) & Text (\%) & All (\%) \\
\midrule
CatBoost & $90.0 \pm 5.9$ & $93.0 \pm 5.0$ & \textbf{$91.5 \pm 5.5$} \\
LightGBM & $84.0 \pm 7.2$ & $93.0 \pm 5.0$ & \textbf{$88.5 \pm 6.2$} \\
RF & $85.0 \pm 7.0$ & $90.0 \pm 5.9$ & \textbf{$87.5 \pm 6.5$} \\
XGBoost & $80.0 \pm 7.8$ & $90.0 \pm 5.9$ & \textbf{$85.0 \pm 6.9$} \\
TabPFNv2 & $77.0 \pm 8.2$ & $84.4 \pm 7.5$ & \textbf{$80.7 \pm 7.9$} \\
TabM & $82.0 \pm 7.5$ & $77.0 \pm 8.2$ & \textbf{$79.5 \pm 7.9$} \\
TabDPT & $70.0 \pm 9.0$ & $87.0 \pm 6.6$ & \textbf{$78.5 \pm 7.9$} \\
RealMLP & $78.0 \pm 8.1$ & $78.0 \pm 8.1$ & \textbf{$78.0 \pm 8.1$} \\
TabSTAR & $76.0 \pm 8.4$ & --- & \textbf{$76.0 \pm 8.4$} \\
TabPFN-2.5 & $77.0 \pm 8.2$ & $73.3 \pm 9.1$ & \textbf{$75.2 \pm 8.7$} \\
ConTextTab & $73.0 \pm 8.7$ & --- & \textbf{$73.0 \pm 8.7$} \\
TabICLv2 & $55.0 \pm 9.8$ & $75.0 \pm 8.5$ & \textbf{$65.0 \pm 9.2$} \\
\bottomrule
\end{tabular}
\end{table}

\paragraph{Per-dataset Results.} Tables~\ref{tab:app_results_image} and~\ref{tab:app_results_text} report per-dataset results for all 20 image-tabular and 20 text-tabular datasets, averaged over all learners that have both Frozen and TAR conditions and 5 random seeds, sorted by TAR gain. Negative $R^2$ scores are clipped before averaging. To test whether the gains from adaptation are significant, we pair each learner-fold run under \textit{Joint TAR} with the same learner and fold under \textit{Joint Frozen}, giving 60 paired observations per image dataset (12 learners $\times$ 5 folds) and up to 50 per text dataset (10 learners $\times$ 5 folds), and apply a one-sided paired $t$-test against the null of no gain. All 40 $p$-values are corrected together as a single family with the Benjamini-Hochberg procedure at $\alpha = 0.05$, and Tables~\ref{tab:app_results_image} and~\ref{tab:app_results_text} report the corrected value per dataset. In total, 37 of the 40 datasets show a significant gain, 18 of 20 on the image side and 19 of 20 on the text side. The three exceptions are Painting Price, CS:GO Skins, and Fake Job Postings; two show a small positive gain that the test cannot separate from noise, and one a small negative one, showing that our curation logic could have been stricter as these datasets do not generalize.

\begin{figure}[!ht]
\centering
\includegraphics[width=0.92\linewidth]{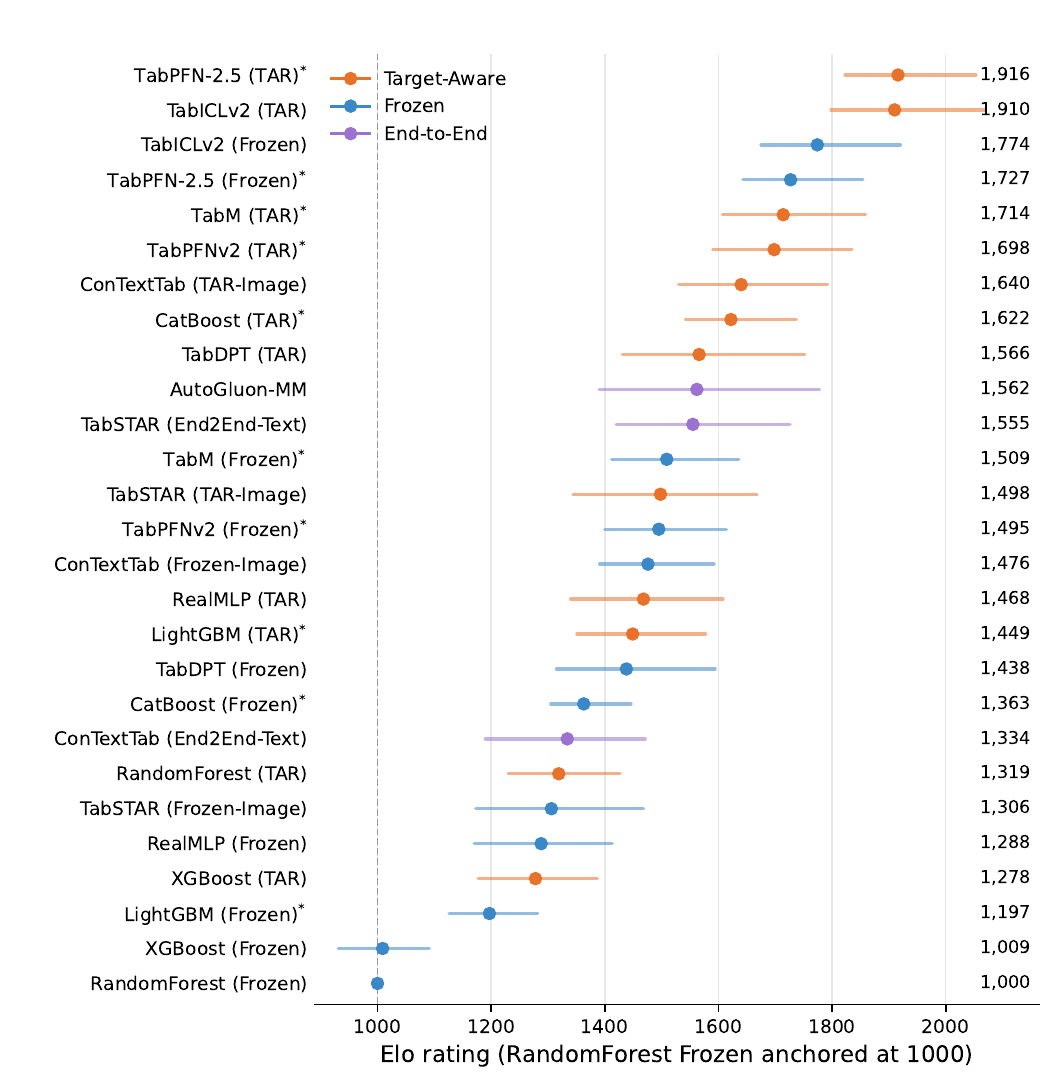}
\caption{Elo over all 40 MulTaBench datasets. Learners marked with $^{*}$ form the curation committee. AutoGluon-MM is a single multimodal system and has no frozen/target-aware split; TabSTAR and ConTextTab are end-to-end models for text, but have frozen and target-aware variants for image.}
\label{fig:elo}
\end{figure}

\paragraph{Elo Leaderboard} Following the protocol adopted by TabArena, we rank competitors using the Bradley-Terry model. We define a battle as the head-to-head comparison of two models on a single dataset-fold, with missing evaluations safely omitted. Ratings are mapped onto the standard Elo scale and anchored by setting RandomForest under \textit{Joint Frozen} to 1000, with confidence intervals derived from dataset-level bootstrap resampling. Figure~\ref{fig:elo} plots each learner by condition, placing \textit{Joint Frozen} and \textit{Joint TAR} in direct competition. Of the 27 total entries, 12 are evaluated under both conditions. Crucially, all 12 models achieve a higher Elo rating with target-aware representations, securing improvements that range from 128 points (TabDPT) to 319 points (RandomForest).

\begin{table*}[ht]
\centering
\caption{MulTaBench Image-Tabular Per-dataset Results. Averaged over 12 learners and 5 seeds, with both Frozen and TAR conditions, sorted by Gain. AUROC for classification, $R^2$ for regression. $p_{\text{BH}}$ is the Benjamini-Hochberg adjusted significance of the gain.}
\label{tab:app_results_image}
\small
\begin{tabular}{lcccc}
\toprule
Dataset & Frozen & TAR & Gain & $p_{\text{BH}}$ \\
\midrule
Mango Mass & 0.533 & 0.653 & $+$0.120 & $<$0.001 \\
Khaadi Clothes & 0.565 & 0.683 & $+$0.118 & $<$0.001 \\
Amazon Packages & 0.523 & 0.579 & $+$0.056 & $<$0.001 \\
CheXpert & 0.762 & 0.803 & $+$0.041 & $<$0.001 \\
CBIS-DDSM & 0.858 & 0.893 & $+$0.035 & $<$0.001 \\
Amazon Bestseller & 0.543 & 0.566 & $+$0.024 & $<$0.001 \\
MkPhoto Bots & 0.341 & 0.361 & $+$0.020 & $<$0.001 \\
Justin Instagram & 0.956 & 0.974 & $+$0.017 & $<$0.001 \\
Hateful Meme & 0.745 & 0.760 & $+$0.015 & $<$0.001 \\
H\&M Fashion & 0.389 & 0.404 & $+$0.015 & $<$0.001 \\
Glaucoma SMDG & 0.921 & 0.934 & $+$0.012 & $<$0.001 \\
Mammography CMMD & 0.770 & 0.780 & $+$0.010 & 0.005 \\
CelebA Attractiveness & 0.903 & 0.913 & $+$0.009 & $<$0.001 \\
PetFinder & 0.832 & 0.841 & $+$0.009 & $<$0.001 \\
HubMAP HPA & 0.959 & 0.968 & $+$0.009 & $<$0.001 \\
Flower Bouquets & 0.627 & 0.636 & $+$0.009 & 0.010 \\
Painting Price & 0.270 & 0.275 & $+$0.005 & 0.236 \\
Letterboxd Movies & 0.439 & 0.443 & $+$0.004 & 0.002 \\
Zooscan Plankton & 0.983 & 0.985 & $+$0.003 & $<$0.001 \\
CS:GO Skins & 0.871 & 0.870 & $-$0.001 & 0.763 \\
\midrule
\textit{Mean} & 0.690 & 0.716 & $+$0.027 & \\
\bottomrule
\end{tabular}
\end{table*}

\begin{table*}[ht]
\centering
\caption{MulTaBench Text-Tabular Per-dataset Results. Averaged over 10 learners and 5 seeds, with both Frozen and TAR conditions, sorted by Gain. AUROC for classification, $R^2$ for regression. $p_{\text{BH}}$ is the Benjamini-Hochberg adjusted significance of the gain.}
\label{tab:app_results_text}
\small
\begin{tabular}{lcccc}
\toprule
Dataset & Frozen & Contextualized & Gain & $p_{\text{BH}}$ \\
\midrule
Jigsaw Toxicity & 0.806 & 0.926 & $+$0.119 & $<$0.001 \\
Video Games Sales & 0.348 & 0.385 & $+$0.036 & 0.004 \\
Mercari & 0.412 & 0.436 & $+$0.024 & $<$0.001 \\
Baby Products & 0.873 & 0.895 & $+$0.022 & $<$0.001 \\
Vancouver Salaries & 0.753 & 0.775 & $+$0.022 & $<$0.001 \\
Rotten Tomatoes & 0.490 & 0.508 & $+$0.018 & $<$0.001 \\
Kickstarter & 0.720 & 0.737 & $+$0.017 & $<$0.001 \\
Book Price & 0.529 & 0.543 & $+$0.013 & 0.001 \\
Zomato Restaurants & 0.818 & 0.832 & $+$0.013 & $<$0.001 \\
Michelin Guide & 0.896 & 0.909 & $+$0.013 & $<$0.001 \\
Book Readability & 0.809 & 0.821 & $+$0.012 & $<$0.001 \\
Wine Review & 0.968 & 0.976 & $+$0.009 & $<$0.001 \\
Product Sentiment & 0.901 & 0.909 & $+$0.009 & $<$0.001 \\
SciMagojr Impact & 0.852 & 0.860 & $+$0.009 & $<$0.001 \\
US Accidents & 0.965 & 0.974 & $+$0.008 & $<$0.001 \\
Women's Clothing & 0.903 & 0.909 & $+$0.006 & $<$0.001 \\
Spotify Genres & 0.935 & 0.940 & $+$0.005 & $<$0.001 \\
Data Scientist Salary & 0.823 & 0.828 & $+$0.005 & $<$0.001 \\
Montgomery Salaries & 0.968 & 0.972 & $+$0.004 & $<$0.001 \\
Fake Job Postings & 0.916 & 0.918 & $+$0.002 & 0.459 \\
\midrule
\textit{Mean} & 0.784 & 0.803 & $+$0.018 & \\
\bottomrule
\end{tabular}
\end{table*}

\subsection{Missing Baselines}

We deliberately exclude autoregressive generative models (LLMs and VLMs) from the benchmark evaluation, due to prohibitive inference costs and a memorization risk. The research on benchmarking LLMs and VLMs for MMTL task still needs to be explored. Although TIME~\cite{luo_time_2025} and MultimodalTabPFN~\cite{kim_multimodalpfn_2025} are relevant baselines, TIME has not released the code at the time of our submission. MultimodalTabPFN, on the contrary, has a working codebase, but it is highly not flexible to serve the model using the popular \textit{scikit-learn} \cite{pedregosa2011scikit} wrapper, making it hard to evaluate.

\subsection{Computation Costs}\label{app:costs}

\begin{figure}[ht]
\centering
\includegraphics[width=0.98\linewidth]{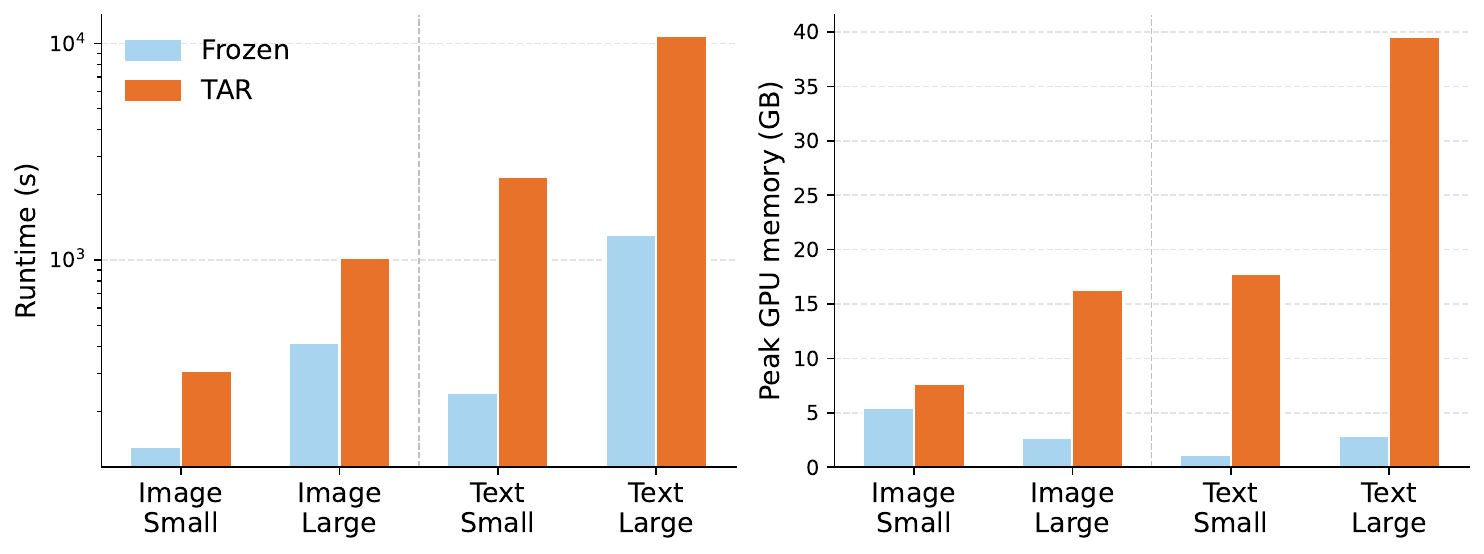}
\caption{Computation costs per run. Left: median runtime in seconds (log scale). Right: median peak GPU memory. The dashed vertical line separates image (left) and text (right) conditions.}
\label{fig:compute_costs}
\end{figure}

Table~\ref{tab:compute_costs} and Figure~\ref{fig:compute_costs} reports median wall-clock runtimes and peak GPU memory per (dataset, fold) run on a single \textit{NVIDIA A100-SXM4} GPU with 40GB memory,
and 8 CPU cores of type \textit{AMD EPYC 7742 64-Core Processor}. We report results for each of the 5 core learners across frozen and TAR conditions and both encoder sizes. From the table, it is evident that the embeddings dominate all metrics. TAR adds a substantial overhead relative to frozen embeddings, dominated by the encoder fine-tuning step. For image datasets with the small DINO encoder, TAR roughly doubles runtime; the large encoder raises costs further.

Text TAR is significantly more expensive: \textit{e5-small} TAR takes roughly ten times longer than frozen, and \textit{e5-large} TAR approaches three hours per run. The gap arises partly as text-tabular datasets often contain more than a single text column, making their effective dataset size much bigger.

The costs above are measured without any hyperparameter optimization (HPO).
Standardizing HPO across 40 datasets is computationally prohibitive under the TAR paradigm: the encoder must be fine-tuned separately for each cross-validation fold to prevent data leakage, so a standard HPO sweep would require repeating encoder fine-tuning for every hyperparameter trial, multiplying an already expensive operation by the number of trials.
Consequently, all experiments use a single fixed LoRA configuration across all datasets, with no per-dataset tuning of the encoder or the learner.
All reported gains should therefore be interpreted as conservative lower bounds on what a fully tuned system could achieve. 

\begin{table*}[ht]
\centering
\caption{Computation costs per run. Median Runtime in seconds and Median Peak GPU memory in GB. Partitioned by tabular learners, modality and encoder size.}
\label{tab:compute_costs}
\setlength{\tabcolsep}{4pt}\small
\begin{tabular}{l rr rr rr rr}
\toprule
& \multicolumn{4}{c}{\textit{Small Encoder}} & \multicolumn{4}{c}{\textit{Large Encoder}} \\
\cmidrule(lr){2-5}\cmidrule(lr){6-9}
& \multicolumn{2}{c}{Runtime (s)} & \multicolumn{2}{c}{Peak GPU (GB)} & \multicolumn{2}{c}{Runtime (s)} & \multicolumn{2}{c}{Peak GPU (GB)} \\
\cmidrule(lr){2-3}\cmidrule(lr){4-5}\cmidrule(lr){6-7}\cmidrule(lr){8-9}
\textbf{Model} & Frozen & TAR & Frozen & TAR & Frozen & TAR & Frozen & TAR \\
\midrule
\multicolumn{9}{l}{\textit{Image-Tabular (DINO encoder)}} \\
\midrule
LightGBM & 141 & 287 & 5.5 & 7.6 & 411 & 1,003 & 2.7 & 16.7 \\
CatBoost & 128 & 307 & 5.4 & 7.6 & 432 & 1,002 & 2.7 & 16.7 \\
TabM & 129 & 282 & 5.4 & 7.6 & 375 & 993 & 2.7 & 16.8 \\
TabPFNv2 & 173 & 318 & 5.3 & 7.6 & 432 & 1,055 & 2.7 & 14.8 \\
TabPFN-2.5 & 169 & 316 & 5.5 & 7.6 & 405 & 1,048 & 2.7 & 14.8 \\
\midrule
\multicolumn{9}{l}{\textit{Text-Tabular (E5 encoder)}} \\
\midrule
LightGBM & 223 & 2,417 & 1.1 & 17.3 & 1,284 & 10,867 & 2.9 & 39.6 \\
CatBoost & 323 & 2,471 & 1.1 & 17.3 & 1,377 & 10,860 & 2.9 & 39.6 \\
TabM & 225 & 2,408 & 1.2 & 17.3 & 1,256 & 10,833 & 2.9 & 39.6 \\
TabPFNv2 & 253 & 2,430 & 1.1 & 18.6 & 1,368 & 11,569 & 2.9 & 39.6 \\
TabPFN-2.5 & 242 & 2,396 & 1.1 & 18.6 & 1,347 & 11,556 & 2.9 & 39.6 \\
\bottomrule
\end{tabular}
\end{table*}

\subsection{Encoder Scale by Task Type}
\label{app:extended_analysis}

\begin{figure*}[ht]
    \centering
    \includegraphics[width=0.98\textwidth]{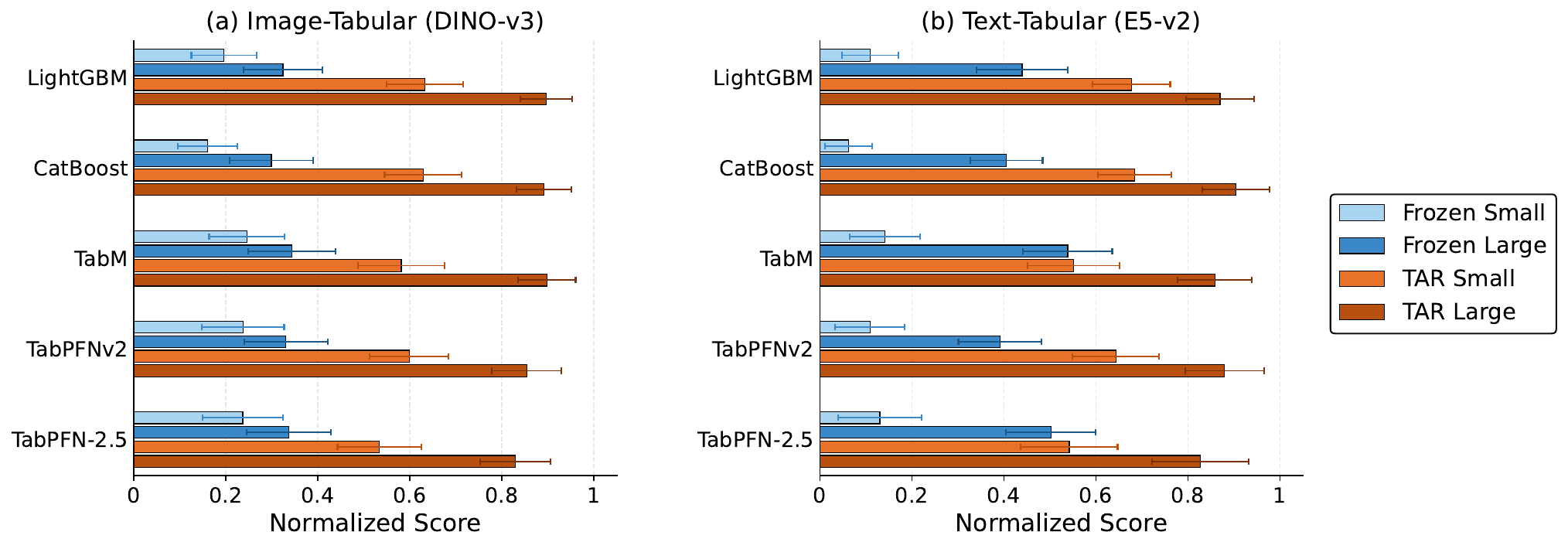}
    \caption{Encoder Scale Analysis for Classification. Small and large encoder variants, frozen and TAR, normalized within each model.}
    \label{fig:encoder_scale_cls}
\end{figure*}

\begin{figure*}[!htb]
    \centering
    \includegraphics[width=0.98\textwidth]{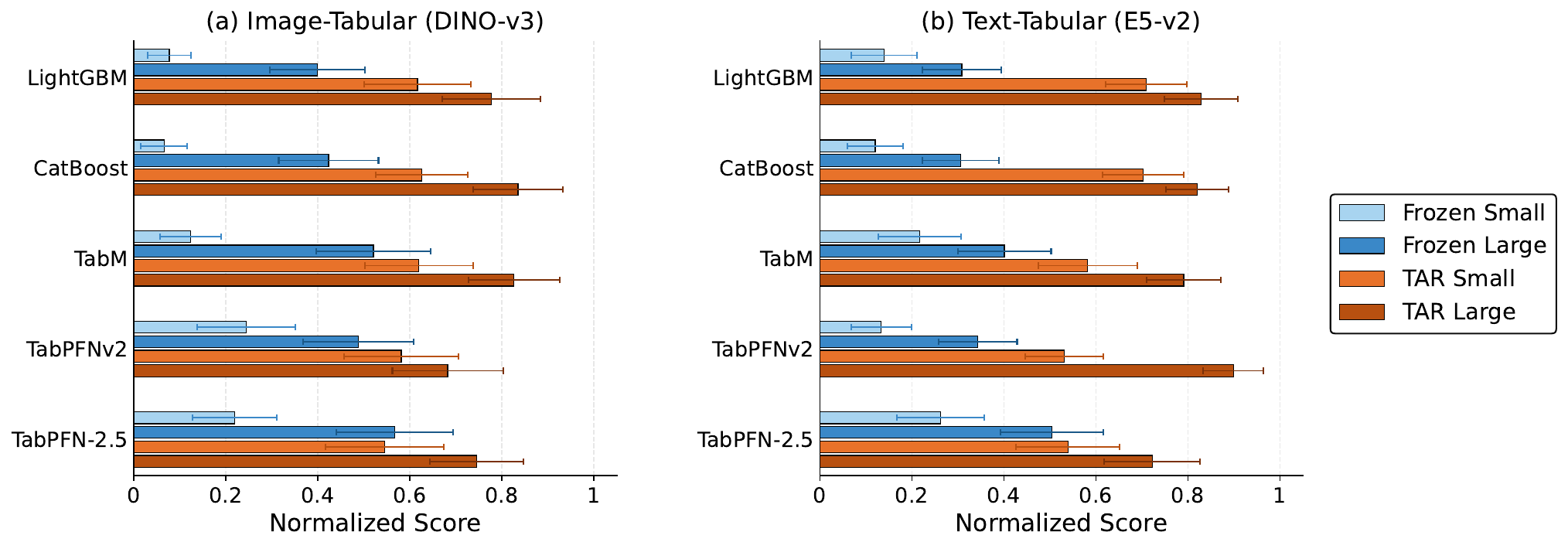}
    \caption{Encoder Scale Analysis for Regression. Small and large encoder variants, frozen and TAR, normalized within each model.}
    \label{fig:encoder_scale_reg}
\end{figure*}

We replace our default encoders, DINO-small and E5-small, with DINO-large\footnote{The official names are \href{https://huggingface.co/facebook/dinov3-vits16-pretrain-lvd1689m}{\textit{dinov3-vits16-pretrain-lvd1689m}} for small, and \href{https://huggingface.co/facebook/dinov3-vitl16-pretrain-lvd1689m}{\textit{dinov3-vitl16-pretrain-lvd1689m}} for large.} and e5-large. Roughly speaking, this moves from models of 30M paremeters to models of 300M parameters. We then re-evaluate all 20 image and 20 text datasets.
Figures~\ref{fig:encoder_scale_cls} and~\ref{fig:encoder_scale_reg} replicate Figure~\ref{fig:encoder_scale} restricted to classification and regression datasets respectively.
TAR consistently outperforms Frozen across encoder sizes and task types, confirming that the benefit of TAR generalizes also to larger encoders.

\subsection{No PCA Variant}
\label{app:extended_analysis_no_pca}

\begin{figure}[ht]
\centering
\includegraphics[width=0.65\linewidth]{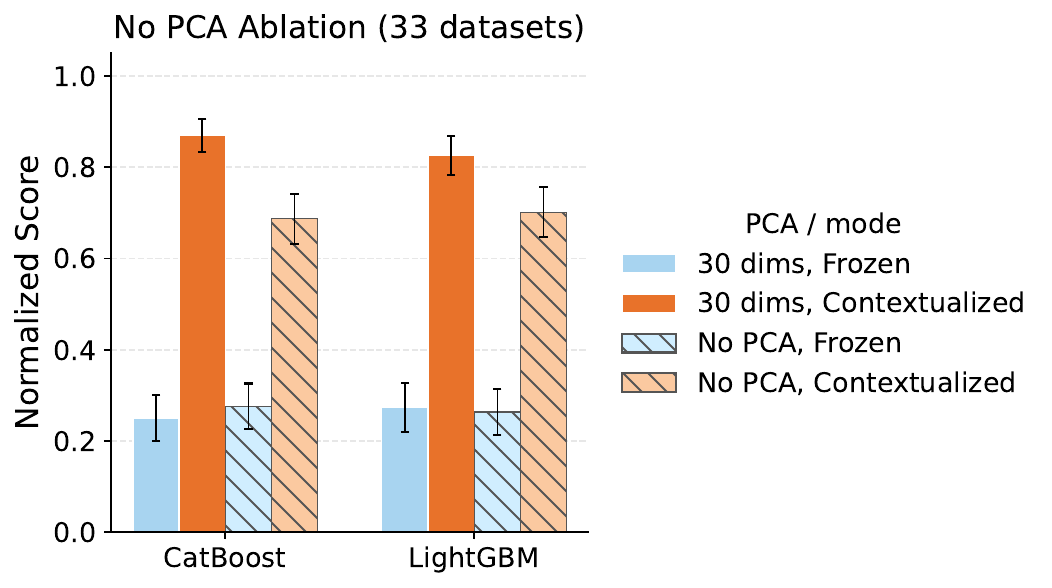}
\caption{No-PCA ablation on 33 datasets for CatBoost and LightGBM. Normalized scores are on the model level.}
\label{fig:nopca}
\end{figure}

To verify that the gains from target-aware adaptation do not depend on the PCA compression step, we repeat the core Frozen vs.\ TAR comparison using raw 384-dimensional embeddings, omitting the projection entirely. Since this largely increases the number of features for the downstream task, we limit the analysis to CatBoost and LightGBM, and exclude datasets with more than 5 text features, resulting in 33 datasets. Figure~\ref{fig:nopca} shows 4 conditions side by side, varying between N=30 and No-PCA, and Frozen vs TAR. We observe that TAR outperforms Frozen in both settings, for both learners, confirming that the advantage is not an artifact of dimensionality reduction.
The signal surfaced by fine-tuning is present in the raw 384-dimensional space and persists regardless of whether embeddings are subsequently compressed.

\subsection{Curation Committee}\label{app:curation_committee}

\begin{figure}[h]
\centering
\includegraphics[width=0.8\linewidth]{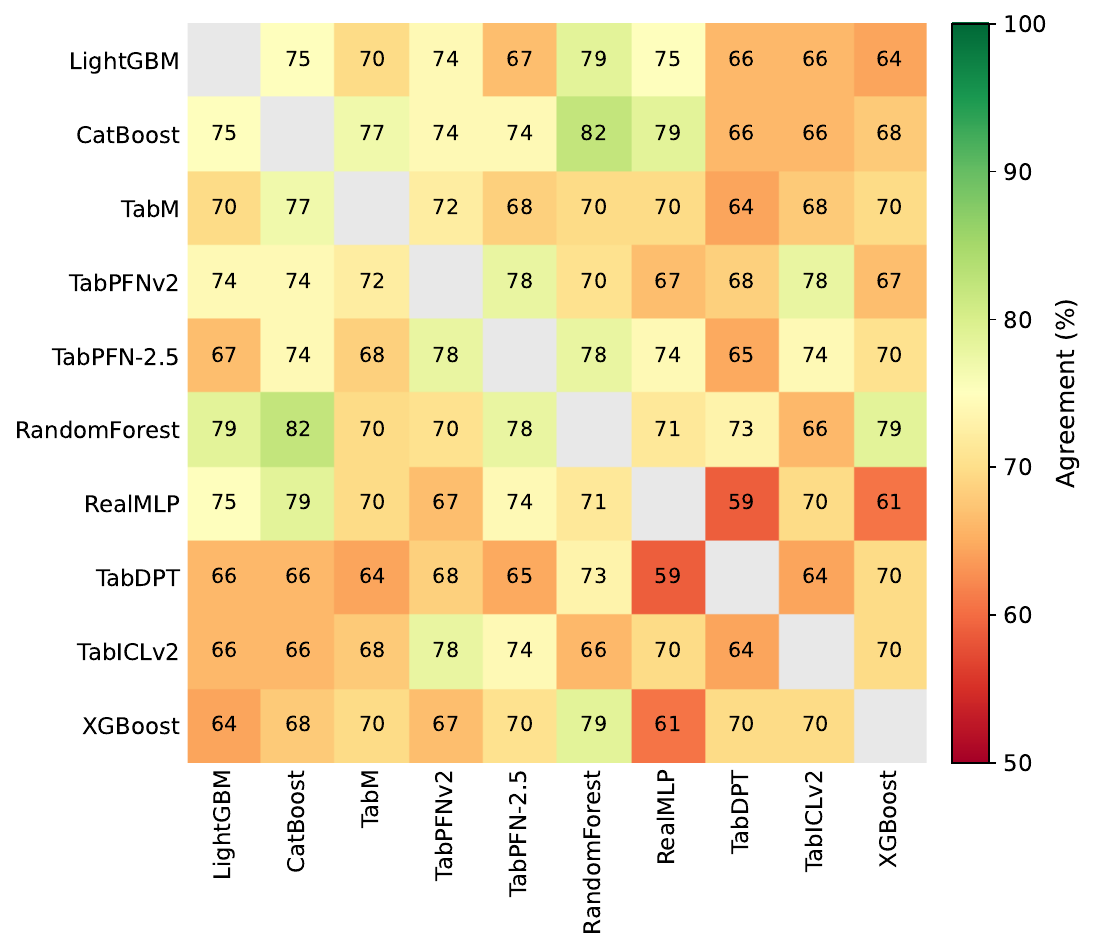}
\caption{Pairwise agreement of the 10 pool learners on the 56-dataset text-tabular pool.}
\label{fig:model_agreement}
\end{figure}

A benchmark curated by a fixed panel of learners risks encoding the algorithmic preferences of that specific committee. To investigate this, we test the acceptance rule under alternative learner panels using the 56-dataset text-tabular pool. We simulate acceptance for every 5-learner panel that can be drawn from the 10, which are $\binom{10}{5} = 252$ options.

\begin{table}[ht]
\centering
\caption{Per-candidate acceptance rate across every 5-learner committee, on the 56-dataset text-tabular pool at $\delta = 0.001$ and $\rho = 3/5$. \textit{Decision}: the decision the published panel reached, where $\checkmark$ is accepted and released, $(\checkmark)$ accepted but held out to match the image subset's size, and $\times$ rejected. \textit{Rate}: share of the $\binom{10}{5} = 252$ committees drawable from the 10-learner pool that admit the dataset.}
\label{tab:panel_pass_rates}
\scriptsize
\setlength{\tabcolsep}{4pt}
\begin{tabular}{lcr@{\hskip 16pt}lcr}
\toprule
Dataset & Decision & Rate & Dataset & Decision & Rate \\
\midrule
Baby Products & $\checkmark$ & 100.0\% & Used Cars Saudi & $\times$ & 26.2\% \\
Book Readability & $\checkmark$ & 100.0\% & WikiLiq Spirits & $\times$ & 26.2\% \\
Data Scientist Salary & $\checkmark$ & 100.0\% & AI Job Salaries & $\times$ & 8.3\% \\
Jigsaw Toxicity & $\checkmark$ & 100.0\% & BikeWale Bikes & $\times$ & 8.3\% \\
Kickstarter & $\checkmark$ & 100.0\% & California Prices & $\times$ & 8.3\% \\
Mercari Marketplace & $\checkmark$ & 100.0\% & JC Penney & $\times$ & 8.3\% \\
Michelin Guide & $\checkmark$ & 100.0\% & Korean Drama & $\times$ & 8.3\% \\
Product Sentiment & $\checkmark$ & 100.0\% & SF Permits & $\times$ & 8.3\% \\
Rotten Tomatoes & $\checkmark$ & 100.0\% & US Museums & $\times$ & 8.3\% \\
SciMagojr Impact & $\checkmark$ & 100.0\% & Used Cars Mercedes & $\times$ & 8.3\% \\
Spotify Genres & $\checkmark$ & 100.0\% & Used Cars Pakistan & $\times$ & 8.3\% \\
US Accidents & $\checkmark$ & 100.0\% & Wine Poland & $\times$ & 8.3\% \\
Vancouver Salaries & $\checkmark$ & 100.0\% & Airbnb Melbourne & $\times$ & 0.0\% \\
Video Games Sales & $\checkmark$ & 100.0\% & Airbnb Seattle & $\times$ & 0.0\% \\
Wine Review & $\checkmark$ & 100.0\% & Anime Planet & $\times$ & 0.0\% \\
Women's Clothing & $\checkmark$ & 100.0\% & Beer Ratings & $\times$ & 0.0\% \\
Zomato Restaurants & $\checkmark$ & 100.0\% & Box Office & $\times$ & 0.0\% \\
Book Price & $\checkmark$ & 91.7\% & CarDekho Cars & $\times$ & 0.0\% \\
Hearthstone Cards & $(\checkmark)$ & 91.7\% & Coffee Reviews & $\times$ & 0.0\% \\
Montgomery Salaries & $\checkmark$ & 73.8\% & Company Size & $\times$ & 0.0\% \\
News Channel & $(\checkmark)$ & 73.8\% & FilmTV Ratings & $\times$ & 0.0\% \\
Consumer Complaints & $(\checkmark)$ & 50.0\% & Goodreads Books & $\times$ & 0.0\% \\
FIFA22 Wages & $\times$ & 50.0\% & Google QA & $\times$ & 0.0\% \\
Fake Job Postings & $\checkmark$ & 50.0\% & IMDB Genre & $\times$ & 0.0\% \\
Wine Vivino Spain & $\times$ & 50.0\% & Indian Laptops & $\times$ & 0.0\% \\
American Eagle & $\times$ & 26.2\% & NBA Draft Value & $\times$ & 0.0\% \\
Chocolate Bars & $\times$ & 26.2\% & Ramen Ratings & $\times$ & 0.0\% \\
OSHA Injury & $\times$ & 26.2\% & Yelp Reviews & $\times$ & 0.0\% \\
\bottomrule
\end{tabular}
\end{table}

Table~\ref{tab:panel_pass_rates} details the acceptance rate for every candidate across all 252 simulated committees. The results strongly support the robustness of the selection process, naturally grouping the candidate datasets into four tiers of multimodal signal strength:
\begin{itemize}
\item \textbf{Full Consensus:} 33 datasets generated complete consensus, consistently passing (17 datasets) or failing (16 datasets) across 100\% of all committee combinations.
\item \textbf{Near Consensus:} 12 datasets exhibited near-unanimous agreement. The lowest accepted dataset in this tier passed in 91.7\% of committees, while the highest rejected dataset passed in only 8.3\%.
\item \textbf{Strong Majority:} 7 datasets maintained a clear and stable majority. This tier included 2 accepted datasets with a 73.8\% acceptance rate and 5 rejected datasets with a 26.2\% acceptance rate.
\item \textbf{Borderline Candidates:} The remaining 4 datasets sat exactly on the decision boundary with a 50.0\% acceptance rate.
\end{itemize}

Given that two of the five models are from the TabPFN family, we also investigate whether their shared architecture acts as a voting bloc that biases the overall rates. Figure~\ref{fig:model_agreement} measures the pairwise agreement across all 10 pool learners. We find that while TabPFNv2 and TabPFN-2.5 are positively correlated, their agreement rate (78\%) is identical to that of completely distinct model families, such as RandomForest and TabPFN-2.5. The average pairwise agreement across all models is 70\%, ranging from 59\% (TabDPT and RealMLP) to 82\% (CatBoost and RandomForest). This healthy variance confirms that the two TabPFN variants do not act as an isolated voting bloc. Ultimately, these results provide strong evidence that our dataset selection is highly robust to the specific choice of tabular learners.

\subsection{Threshold Sensitivity}\label{app:threshold_sensitivity}

The acceptance rule elaborated in Appendix~\ref{app:curation_formal} has two hyperparameters: a minimal performance margin ($\delta = 0.001$) and a simple majority vote ($\rho = 3/5$) to generate consensus across learners. While this design imposes a strict binary cutoff for inclusion, the underlying demand for target-awareness naturally exists on a spectrum. Some datasets cross the threshold marginally, whereas others generate strong, undeniable consensus across learners. In this subsection, we analyze how the benchmark varies when demanding stricter thresholds.

\begin{table}[ht]
\centering
\caption{\textbf{Threshold sensitivity analysis.} Number of datasets (out of 40) retained when varying the required performance margin ($\delta$) and learner consensus ($\rho$). The baseline setting is in bold.}
\label{tab:threshold_grid}
\small
\begin{tabular}{lccc}
\toprule
& \multicolumn{3}{c}{Consensus ($\rho$)} \\
\cmidrule(lr){2-4}
Margin ($\delta$) & \textbf{3/5} & $4/5$ & $5/5$ \\
\midrule
\textbf{0.001} & \textbf{40} & 33 & 23 \\
0.002 & 36 & 30 & 20 \\
0.005 & 30 & 21 & 13 \\
0.01 & 21 & 12 & 2 \\
0.02 & 7 & 3 & 2 \\
\bottomrule
\end{tabular}
\end{table}

Table~\ref{tab:threshold_grid} details the number of datasets retained across different combinations of $\delta$ and $\rho$. Elevating the consensus requirement from a simple majority ($3/5$) to $4/5$ retains 33 datasets, naturally discarding those that marginally cleared the initial curation. Even under strict unanimity ($5/5$), 23 datasets survive, demonstrating that MulTaBench datasets generate broad agreement across tabular models.

Similarly, increasing the performance margin $\delta$ systematically isolates the strongest target-aware signals. Doubling the margin to $0.002$ retains 36 datasets, and raising it to $0.005$ retains 30. Furthermore, we observe that demanding a strict $1\%$ absolute improvement ($\delta = 0.01$) still preserves over half the benchmark (21 datasets). This is a notably high bar in tabular learning, where performance gaps between competitive models are often narrower. Ultimately, this analysis confirms that the need for target-aware representations falls along a continuous spectrum, and that applying stricter thresholds reliably acts as a filter to surface the most pronounced multimodal tasks.

\section{Additional Attention Maps}\label{app:attention_appendix}

Each of the 4 datasets in Figure~\ref{fig:attention_main} is accompanied by 3 additional test-set examples below. In every case, Frozen attention remains scattered across task-irrelevant regions, while Target-Aware attention converges on semantically meaningful area identified in the main figure, relevant to the prediction.

\begin{figure}[ht]
\centering
\setlength{\tabcolsep}{1pt}
\begin{tabular}{ccc}
\small\textbf{Image} & \small\textbf{Frozen} & \small\textbf{Target-Aware} \\[2pt]
\includegraphics[width=0.31\linewidth]{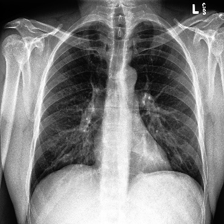} &
\includegraphics[width=0.31\linewidth]{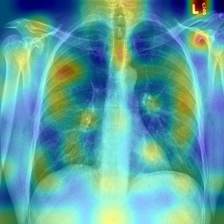} &
\includegraphics[width=0.31\linewidth]{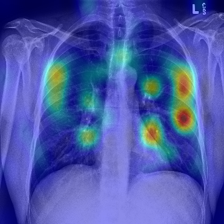} \\
\includegraphics[width=0.31\linewidth]{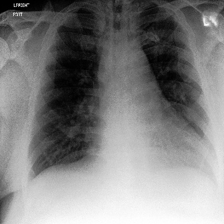} &
\includegraphics[width=0.31\linewidth]{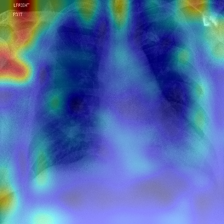} &
\includegraphics[width=0.31\linewidth]{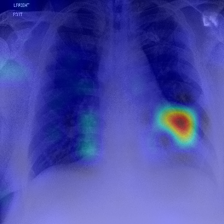} \\
\includegraphics[width=0.31\linewidth]{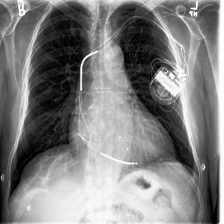} &
\includegraphics[width=0.31\linewidth]{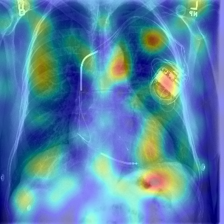} &
\includegraphics[width=0.31\linewidth]{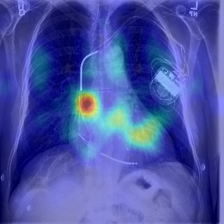} \\
\end{tabular}
\caption{CheXpert Attention Maps. The attention shifts from diffused edges to the lung.}
\label{fig:attn_chexpert_appendix}
\end{figure}

\begin{figure}[ht]
\centering
\setlength{\tabcolsep}{1pt}
\begin{tabular}{ccc}
\small\textbf{Image} & \small\textbf{Frozen} & \small\textbf{Target-Aware} \\[2pt]
\includegraphics[width=0.31\linewidth]{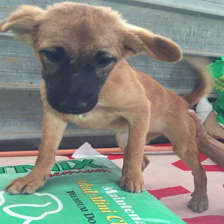} &
\includegraphics[width=0.31\linewidth]{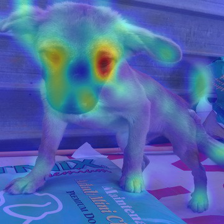} &
\includegraphics[width=0.31\linewidth]{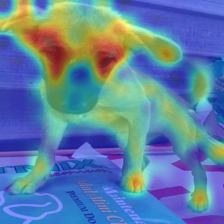} \\
\includegraphics[width=0.31\linewidth]{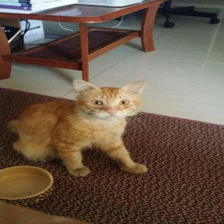} &
\includegraphics[width=0.31\linewidth]{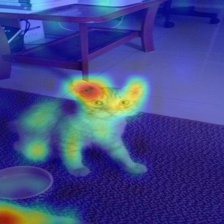} &
\includegraphics[width=0.31\linewidth]{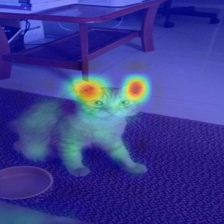} \\
\includegraphics[width=0.31\linewidth]{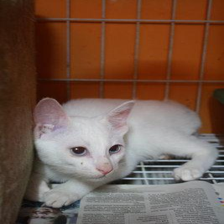} &
\includegraphics[width=0.31\linewidth]{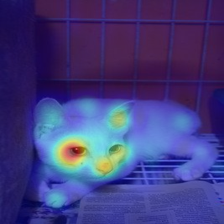} &
\includegraphics[width=0.31\linewidth]{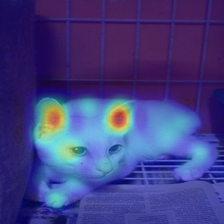} \\
\end{tabular}
\caption{PetFinder Attention Maps. Attention isolates the cat ears and the dog's eyes.}
\label{fig:attn_petfinder_appendix}
\end{figure}

\begin{figure}[ht]
\centering
\setlength{\tabcolsep}{1pt}
\begin{tabular}{ccc}
\small\textbf{Image} & \small\textbf{Frozen} & \small\textbf{Target-Aware} \\[2pt]
\includegraphics[width=0.31\linewidth]{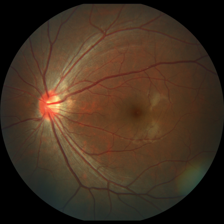} &
\includegraphics[width=0.31\linewidth]{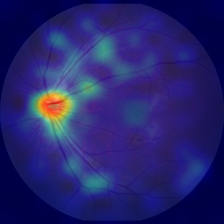} &
\includegraphics[width=0.31\linewidth]{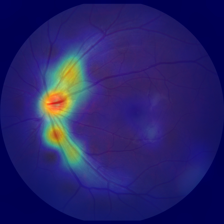} \\
\includegraphics[width=0.31\linewidth]{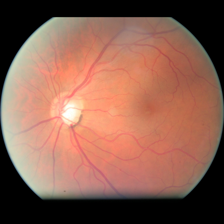} &
\includegraphics[width=0.31\linewidth]{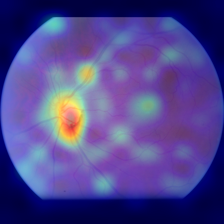} &
\includegraphics[width=0.31\linewidth]{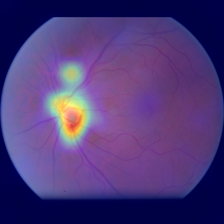} \\
\includegraphics[width=0.31\linewidth]{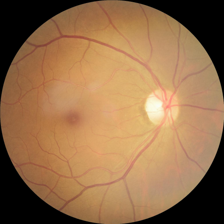} &
\includegraphics[width=0.31\linewidth]{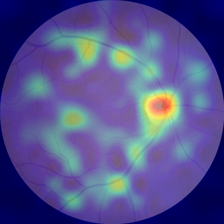} &
\includegraphics[width=0.31\linewidth]{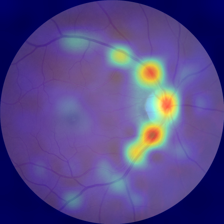} \\
\end{tabular}
\caption{Glaucoma Attention Maps. Frozen attention scatters randomly across the retina; TAR converges on the optic disc and nerve fiber region, the clinically relevant area for glaucoma diagnosis.}
\label{fig:attn_glaucoma_appendix}
\end{figure}

\begin{figure}[ht]
\centering
\setlength{\tabcolsep}{1pt}
\begin{tabular}{ccc}
\small\textbf{Image} & \small\textbf{Frozen} & \small\textbf{Target-Aware} \\[2pt]
\includegraphics[width=0.31\linewidth]{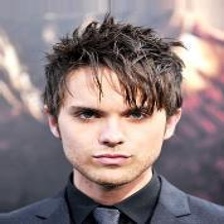} &
\includegraphics[width=0.31\linewidth]{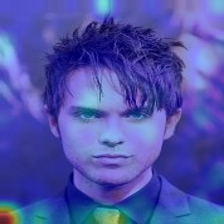} &
\includegraphics[width=0.31\linewidth]{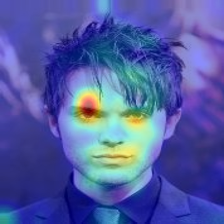} \\
\includegraphics[width=0.31\linewidth]{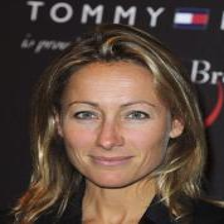} &
\includegraphics[width=0.31\linewidth]{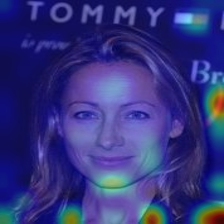} &
\includegraphics[width=0.31\linewidth]{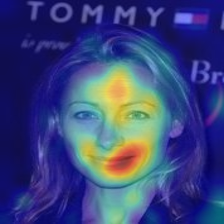} \\
\includegraphics[width=0.31\linewidth]{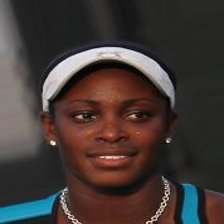} &
\includegraphics[width=0.31\linewidth]{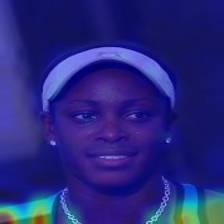} &
\includegraphics[width=0.31\linewidth]{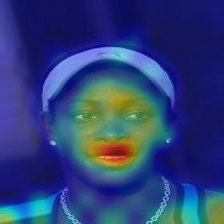} \\
\end{tabular}
\caption{Celeb Attractiveness Attention Maps. Frozen attention disperses across accessories, clothing, and background; TAR consistently focuses on facial features.}
\label{fig:attn_celeb_appendix}
\end{figure}

\newpage
\clearpage


\end{document}